\documentclass[lettersize,journal]{IEEEtran}
\usepackage{amsmath,amsfonts}
\usepackage{algorithmic}
\usepackage{algorithm}
\usepackage{array}
\usepackage[caption=false,font=normalsize,labelfont=sf,textfont=sf]{subfig}
\usepackage{textcomp}
\usepackage{stfloats}
\usepackage{amssymb,amsmath}
\usepackage{url}
\usepackage{hyperref}
\usepackage{multirow}
\usepackage{verbatim}
\usepackage{graphicx}
\usepackage{cite}
\usepackage[table]{xcolor}
\newcommand{\yzd}[1]{\textcolor{red}{#1}}

\begin{document}

\title{SED-MVS: Segmentation-Driven and Edge-Aligned Deformation Multi-View Stereo with Depth Restoration and Occlusion Constraint}


%

\author{Zhenlong Yuan, Zhidong Yang, Yujun Cai, Kuangxin Wu, Mufan Liu, Dapeng Zhang, Hao Jiang, Zhaoxin Li, and Zhaoqi Wang~\IEEEmembership{Member,~IEEE,}
\thanks{

Zhenlong Yuan, Zhidong Yang, Hao Jiang and Zhaoqi Wang are with the Institute of Computing Technology, Chinese Academy of Sciences, Beijing 100190, China. (e-mail: yuanzhenlong21b@ict.ac.cn, yangzhidong19s@ict.ac.cn, jianghao@ict.ac.cn, zqwang@ict.ac.cn)

Yujun Cai is with School of Electrical Engineering and Computer Science, The University of Queensland (UQ), Australia. (e-mail: yujun.cai@uq.edu.au)

Kuangxin Wu is with the Information Technology Department, Hunan Police Academy, Changsha 410100, China. (e-mail: kuangxinwu@outlook.com)

Mufan Liu is with Cooperative MediaNet Innovation Center, Shanghai Jiao Tong University, Shanghai, 200240, China. (e-mail: sudo\_evan@sjtu.edu.cn)

Dapeng Zhang is with DSLAB, School of Information Science \& Engineering, Lanzhou University, 730000, China. (zhangdp22@lzu.edu.cn)

Zhaoxin Li is with Agricultural Information Institute, Chinese Academy of Agricultural Sciences and Key Laboratory of Agricultural Big Data, Ministry of Agriculture and Rural Affairs, 100081, China. (e-mail: cszli@hotmail.com)





}
}

\markboth{}%
{Shell \MakeLowercase{\textit{et al.}}: A Sample Article Using IEEEtran.cls for IEEE Journals}


\maketitle

\begin{abstract}
Recently, patch-deformation methods have exhibited significant effectiveness in multi-view stereo owing to the deformable and expandable patches in reconstructing textureless areas.
However, such methods primarily emphasize broadening the receptive field in textureless areas, while neglecting deformation instability caused by easily overlooked edge-skipping, potentially leading to matching distortions. 
To address this, we propose SED-MVS, which adopts panoptic segmentation and multi-trajectory diffusion strategy for segmentation-driven and edge-aligned patch deformation. 
Specifically, to prevent unanticipated edge-skipping, we first employ SAM2 for panoptic segmentation as depth-edge guidance to guide patch deformation, followed by multi-trajectory diffusion strategy to ensure patches are comprehensively aligned with depth edges.
Moreover, to avoid potential inaccuracy of random initialization, we combine both sparse points from LoFTR and monocular depth map from DepthAnything V2 to restore reliable and realistic depth map for initialization and supervised guidance.
Finally, we integrate segmentation image with monocular depth map to exploit inter-instance occlusion relationship, then further regard them as occlusion map to implement two distinct edge constraint, thereby facilitating occlusion-aware patch deformation.
Extensive results on ETH3D, Tanks \& Temples, BlendedMVS and Strecha datasets validate the state-of-the-art performance and robust generalization capability of our proposed method.

\end{abstract}



\section{Introduction}
\IEEEPARstart{M}{ulti-view} Stereo (MVS) is a fundamental task in computer vision, aiming to reconstruct the dense 3D structure of the scene or object by utilizing multiple images captured from different viewpoints. Extensive applications of MVS span across heritage preservation, autonomous driving, and medical imaging, leading to the emergence of numerous datasets \cite{ETH3D, TNT, Blendedmvs, strecha} and impressive methodologies \cite{Pm-Huber, COLMAP, Accurate}. Despite these advancements, MVS still suffers from various issues including occlusion, shadows, textureless areas, lambertian surfaces, etc. Such issues have hindered the efficiency and quality of the reconstructed scene or object.

\begin{figure}
\centering
\includegraphics[width=0.95\linewidth]{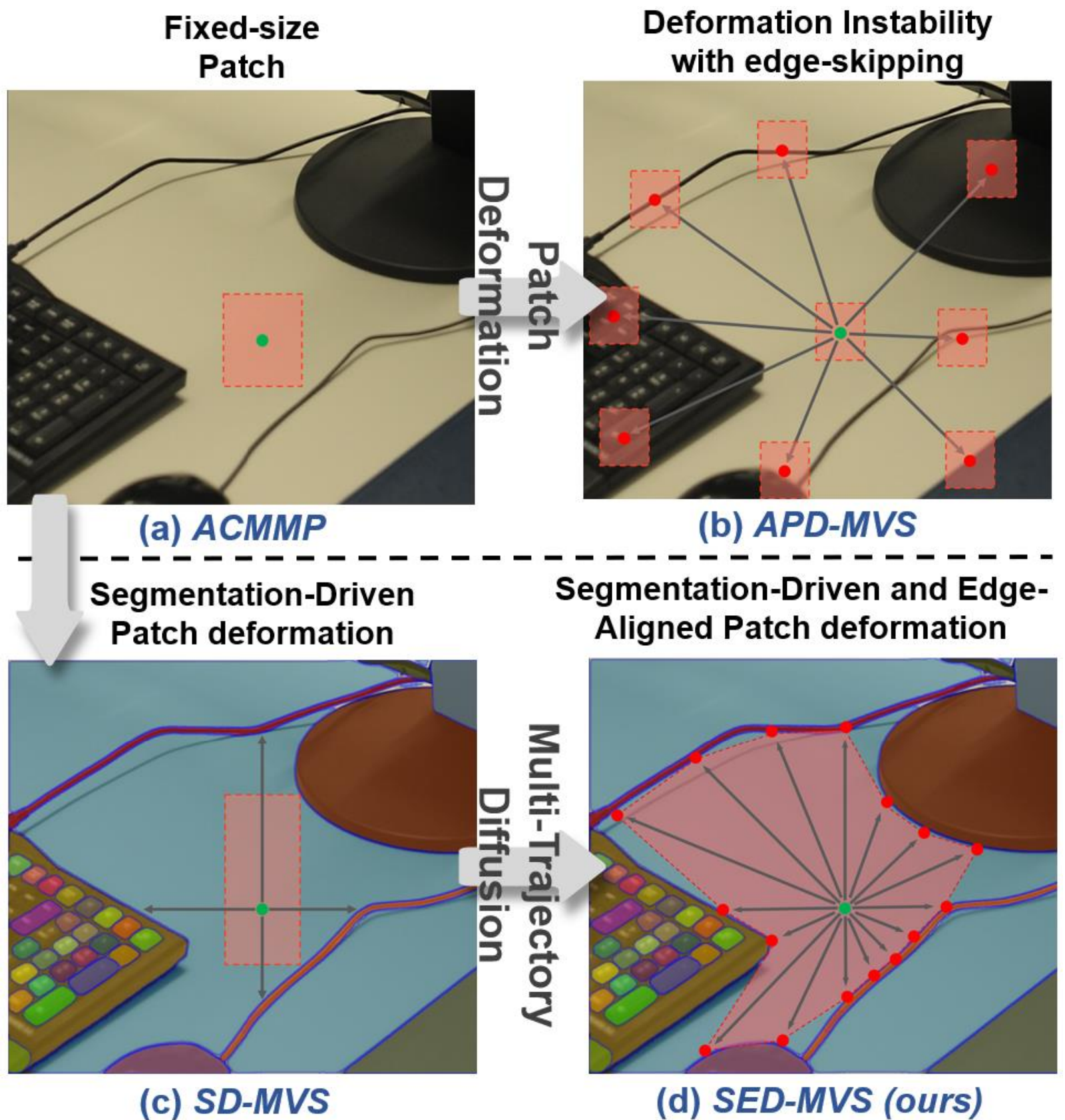}
\caption{
Comparative analysis between ACMMP, APD-MVS, SD-MVS and our method. 
Green and red dots denote the central pixel and boundary pixel, respectively, while the red background indicates the patch.
ACMMP in (a) struggles to reconstruct textureless areas as its fixed-size patch fails to capture sufficient feature points.
The ignorance of depth edge causes deformed patch of APD-MVS in (b) to be selected in depth-discontinuous areas, leading to potential matching distortions.
SD-MVS in (c) adjusts the patch scale based on the distance from pixel to the depth edge boundary, while it remains constrained by the fixed patch size and lacks edge alignment capabilities.
In contrast, our method in (d) adopts panoptic segmentation and multi-trajectory diffusion to enable segmentation-driven and edge-aligned patch deformation.}
\vspace{-0.1in}
\label{fig: Compare}
\end{figure}

To achieve efficient and high-quality 3D reconstruction, various MVS methods have been proposed, which primarily fall into two categories: learning-based MVS and traditional MVS. 
Learning-based MVS utilizes convolutional neural networks to extract high-dimensional representations of 3D cost volumes for reconstruction. Such methods frequently encounter difficulties like unaffordable memory usage or limited generalization capabilities.
In contrast, traditional MVS originated from the PatchMatch (PM) algorithm \cite{PM}, which constructs the solution space by propagating neighboring hypotheses and generating refined hypotheses, then selects the optimal ones through a criterion named multi-view matching costs. 
However, when faced with scenarios characterized by textureless areas, its matching cost inevitably becomes unreliable due to the lack of distinguishable features within the receptive field of its patch.

To reconstruct textureless areas, three primary techniques of traditional MVS have been proposed: coarse-to-fine, planarization and patch deformation strategy. 
Coarse-to-fine strategy \cite{ACMM, MG-MVS, Pyramid} adopts a pyramid architecture to progressively processing multi-scale images from coarse to fine, thus allowing patches to extract multi-scale features for reconstruction. However, limited by the capability of effectively preserving high-level information via pyramid architecture, the coarse-to-fine scheme struggles to reconstruct large texture-less areas.
Moreover, planarization strategy \cite{PCF-MVS, TAPA-MVS, TSAR-MVS, ACMP, HPM-MVS} concatenates reliable pixels in well-textured areas to form connected areas, then regard them as textureless areas and perform planarization for reconstruction. 
Despite various strategies like superpixel \cite{TSAR-MVS}, triangulation \cite{ACMP}, KD-tree \cite{HPM-MVS} have employed, these methods still suffer from the limited size of connected areas and are prone to planarization deviations.

Differently, patch deformation strategy through deformation on fixed-size patches subtly expands the receptive field for reconstruction. 
Without being limited by the number of pyramid layers or the size of connected areas, patch deformation can flexibly handle textureless areas with arbitrary size. 
For instance, API-MVS \cite{API-MVS} and PHI-MVS \cite{PHI-MVS} respectively introduce entropy and dilated convolution on patches to dynamically adjust patch size and sampling intervals. 
Moreover, APD-MVS \cite{APD-MVS} expands outward from each unreliable pixel to identify its correlated reliable pixels, then constructs several sub-patches centered on them to calculate matching costs.

However, these methods mainly focus on advancing reliable pixel search strategies to minimize matching ambiguity, while neglecting the crucial premise of deformation stability. 
As shown in Fig. \ref{fig: Compare} (b), centered on the unreliable green pixel, the patch deformation method first divides its surrounding area into fixed-angle sectors, then searches for a reliable red pixel within each sector and construct a sub-patches centered on it to form the deformable PM, which will displace the traditional PM in Fig. \ref{fig: Compare} (a) for reconstruction. 
Nevertheless, the ignorance of depth continuity premise in patches causes its deformed patch mistakenly crossing crucial depth edges and consequently being selected in depth-discontinuous areas, thus leading to potential matching distortions.

To address this, we propose the \textbf{S}egmentation-driven and \textbf{E}dge-aligned Patch \textbf{D}eformation for \textbf{M}ulti-\textbf{V}iew \textbf{S}tereo (SED-MVS). The proposed SED-MVS adopts SAM-based segmentation prior and multi-trajectory diffusion strategy to achieve an accurate MVS reconstruction. As shown in Fig. \ref{fig: Compare} (d), we first employ panoptic segmentation to extract depth edges as guidance for patch deformation, followed by multi-trajectory diffusion to enable patch aligned with depth edges in complicated scenarios. Then we adopt segmentation-driven triangulation and geometry-aware refinement to combine sparse points and monocular depth map, thereby restoring reliable depth map for initialization and supervised guidance. Finally, we integrate segmentation image with monocular depth map to exploit occlusion relationship between instances, then further regard them as occlusion map
for two distinct edge constraints to achieve occlusion-aware patch deformation.

Compared to the prior work \cite{SD-MVS} in Fig. \ref{fig: Compare} (c), we further exploit the edge alignment capability of deformed patches in complex scenes, denoted as SED-MVS, and the reconstruction quality can be further improved. Specifically, the differences are summarized as follows:
1) In this work, we adopt both panoptic segmentation and multi-trajectory diffusion for segmentation-driven and edge-aligned patch deformation. In contrast, the previous work only adopts panoptic segmentation for patch scaling, which lacks edge alignment capabilities. 
2) We leverage sparse point and monocular depth maps to restore reliable and realistic depth map for initialization before PM process, while the prior work only adopts random initialization.
3) We further exploit inter-instance occlusion relationship for two distinct edge constraint, while the prior work only adopts single edge constraint without occlusion awareness.
4) We conduct more experiments, ablation studies, design analysis, and qualitative results to demonstrate the advantages of our proposed method, and perform comprehensive comparisons with state-of-the-art counterparts on the ETH3D \cite{ETH3D}, Tanks \& Temples \cite{TNT}, BlendedMVS \cite{Blendedmvs} and Strecha \cite{strecha} datasets. 

In summary, we make the following key contributions:
\begin{itemize}
    \item We develop SED-MVS, which leverages panoptic segmentation to extract depth-edge guidance as guidance and propose multi-trajectory diffusion strategy to achieve segmentation-driven and edge-aligned patch deformation.
    \item We propose sparse-monocular synergistic restoration by combining both sparse points and monocular depth maps to restore reliable and realistic depth map for initialization and supervised guidance.
    \item We integrate segmentation image with monocular depth map to obtain occlusion map for dual-categories edge constraint, enabling occlusion-aware patch deformation.
\end{itemize}







\section{Related Work}
\subsection{Traditional MVS Methods}
\subsubsection{PatchMatch MVS}
The basic concept of PatchMatch \cite{PM} targets identifying the optimal patch pairs between images through a pipeline involving random initialization, propagation and refinement. 
PMS \cite{PMS} pioneers to implement this concept in MVS, thus building the foundation for numerous innovative ideas. 
For acceleration, Gipuma \cite{Gipuma} achieves a diffusion-like propagation scheme through the red-black checkerboard, thus enabling algorithm's deployment on GPUs. 
To reconstruct textureless areas, ACMM \cite{ACMM} proposes both adaptive checkerboard scheme and multi-hypothesis joint view selection, which is further improved by ACMMP \cite{ACMMP} which introduces probabilistic graphical model for triangulation planarization. 
Moreover, TAPA-MVS \cite{TAPA-MVS} and PCF-MVS \cite{PCF-MVS} attempt to extract and planarize superpixel for reconstruction, while limited superpixel size cannot adapt to various scenes. 
Furthermore, MG-MVS \cite{MG-MVS} and HPM-MVS \cite{HPM-MVS} leverage a multi-scale pyramid architecture to extract multi-dimensional features for reconstruction.
CLD-MVS \cite{CLD-MVS} and TSAR-MVS \cite{TSAR-MVS} further introduce the confidence-driven estimator to identify unreliable pixels and adopt boundary-aware interpolation for padding. 
Despite notable improvements brought by these algorithms, the core problem of insufficient patch receptive field remains unresolved, thus leaving room for further improvement.
\yzd{}

\subsubsection{Patch Deformation}
As a subset of PatchMatch MVS, patch deformation can adaptively expand the receptive field by adjusting the original fixed-size patch for matching cost. 
For acceleration, PHI-MVS \cite{PHI-MVS} leverages the idea of dilated convolution to dynamically modify its patch size.
API-MVS \cite{PHI-MVS} also proposes the entropy calculation to adaptively change the patch sampling interval.
Moreover, SD-MVS \cite{SD-MVS} adopts instance segmentation to extract depth edge for patch deformation, while such deformation only focuses on scaling the patch rather than adjusting its shape, making it unable to align with depth edges. 
In contrast, APD-MVS \cite{APD-MVS} achieves arbitrary-shaped patch deformation by decomposing each unreliable pixel's patch into several outward-extending sub-patches with high reliability. However, the ignorance of deformation stability causes deformed patches to occur unexpected edge-skipping and accordingly cover the depth discontinuous areas, leading to suboptimal reconstruction.





\begin{figure*}
\centering
\includegraphics[width=\linewidth]{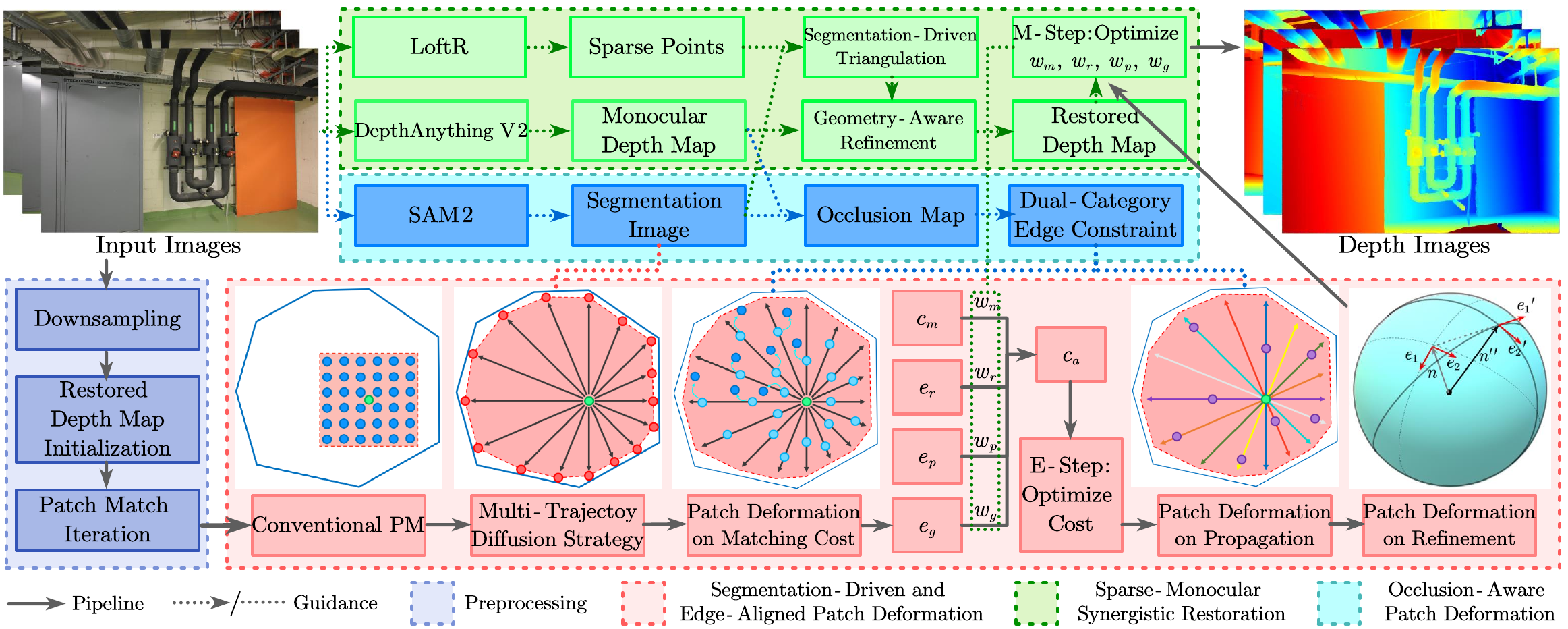}
\caption{
An illustrated pipeline of our method. 
Given the input images, we first adopt LoftR, DepthAnything V2 and SAM2 to obtain their sparse points, monocular depth maps and segmentation images, respectively. Then we regard segmentation images as the depth-edge guidance to accurately constrain the patch deformation within depth-continuous areas, followed by multi-trajectory diffusion to enable edge-aligned patch deformation. The deformed patches are then sequentially processed through texture-aware mapping, load balancing, and spherical gradient refinement to achieve matching cost, propagation, and refinement, respectively.
Moreover, we adopt segmentation-driven triangulation and geometry-aware refinement to combine sparse points and monocular depth map, thereby generating restored depth map for initialization and supervised guidance during the PM process. 
Finally, we integrate segmentation image with monocular depth map to generate occlusion map for dual-category edge constraint, thereby enabling occlusion-aware patch deformation.
} 
\label{fig: pipeline}
\end{figure*}

\subsection{Learning-based MVS Method}
The advancement of deep learning algorithm has led to the development of numerous self-supervised learning methods \cite{chen2023self, chen2024learning, DING1, DING2}, vision-language model \cite{chen2024bimcv, qianmaskfactory, chen2024tokenunify}, parallel computing \cite{Cheng1, Cheng2, Cheng3} transformer-based \cite{SUN1, SUN2} and cluster-based \cite{GUAN1, GUAN2} methods, and compression strategy \cite{Zhang2, Zhang1, Zhang3, Zhang4}.
In MVS field, MVSNet \cite{MVSNet} is the first to leverage convolutional neural networks to build differentiable 3D cost volumes from 2D features, building the foundation for numerous follow-up works.
R-MVSNet \cite{R-MVSNet} and IterMVS-LS \cite{Iter-MVS} further propose the GRU module to regularize cost volumes and encode probability distributions, respectively, thereby facilitating large-scale 3D reconstruction. 
To reduce time and memory usage, Cas-MVSNet \cite{CVP-MVSNet}, CVP-MVSNet \cite{CVP-MVSNet} and UCS-Net \cite{UCS-Net} utilize the feature pyramid to progressively achieve depth estimation in a coarser to fine manner.
Additionally, PatchMatchNet \cite{PatchMatchNet} embeds the idea of PatchMatch into the deep learning framework for acceleration. 
For feature extraction, AAR-MVSNet \cite{Aa-Rmvsnet} present multi-scale aggregation and context-aware convolution to adaptively extract image features. 
Concerning topology, Geo-MVSNet \cite{Geo-MVSNet} and RA-MVSNet \cite{RA-MVSNet} respectively introduce a two-branch geometry fusion network and predict point-to-surface distance from cost volume to enhance geometry and surface topology perception. 
However, most learning-based MVS methods \cite{Su2022, Li2023d, Gu2024, Xu2024, Zhu2024} still struggles with large training datasets, limited generalization capability and unaffordable memory comsumption, making them impractical for large-scale 3D reconstruction.
Therefore, we focus on traditional MVS.

\section{Overview}
Given a series of overlapping images $\mathcal{I} = \{I_i | i = 1, ..., N \}$ and their corresponding camera parameters $\mathcal{P}=\{K_i, R_i, T_i \mid i=1 \cdots N\}$ as input, our goal is to sequentially select the reference image $I_{\mathrm{ref}}$ from all images $\mathcal{I}$, and then adopt multi-view matching with the remaining source images $I_{\mathrm{src}}(\mathcal{I} - I_{\mathrm{ref}})$ to reconstruct its depth map. 
An overview of our pipeline is shown in Fig. \ref{fig: pipeline}, and the design of each proposed component will be detailed in the following sections. 

\section{Method}
\subsection{Segmentation-Driven and Edge-Aligned Patch Deformation}
To reconstruct textureless areas, recent methods attempt to leverage patch deformation to expand the limited patch receptive field, while neglecting the crucial premise of deformation stability. 
As shown in Fig. \ref{fig: Compare} (b), the deformed patch frequently neglects the inconsistent depth and mistakenly covers the areas with discontinuous depth, introducing potential ambiguities for matching cost.
To address this issue, we respectively adopt panoptic segmentation to extract the instance-level boundaries of depth as depth-edge guidance to ensure the patch deformation can accurately fall into the areas with continuous depth. Moreover, we further propose the multi-trajectory diffusion strategy for edge alignment, thereby addressing the ambiguity issue in matching cost and further achieving segmentation-driven and edge-aligned patch deformation.

\begin{figure*}
\centering
\includegraphics[width=\linewidth]{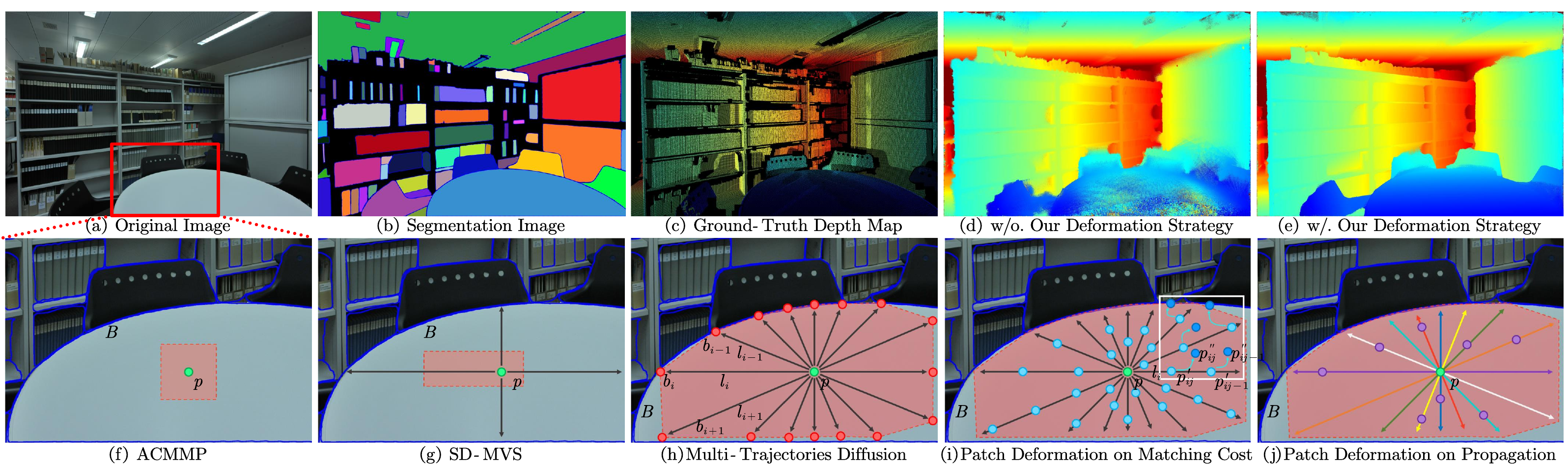}
\caption{
Segmentation-driven and edge-aligned patch deformation. 
In (b), different colors represent different instance masks. 
From (c) to (e), higher color temperatures indicate greater depth values. 
From (f) to (j), instance edges are highlighted in blue, with green dots and red background respectively denoting the center pixels and the deformed patch.
In (g), SD-MVS calculates distances in the vertical and horizontal directions from the center pixel to its boundaries, then proportionally deforms its patch accordingly.
In (h), red dots and black lines respectively indicate boundary pixels and paths.
In (i), cyan and blue dots respectively denote the initially selected pixels and the final mapping pixels for matching cost. For clarity, the texture-aware mapping strategy is only visualized in the white box.
In (j), diagonal paths are grouped and marked with the same color. From each grouped path, one purple point is selected for propagation.
}
\label{fig: Diffusion}
\end{figure*}

\subsubsection{SAM2-based Depth-Edge Guidance}
To fully exploit depth edges, we first introduce SAM2 \cite{SAM2} for panoramic segmentation.
According to \cite{Sengupta2024b}, SAM2 demonstrates superior segmentation performance compared to SAM, especially in scenarios characterized by overlapping instances with extensive similar colors and occlusions like forests and farmlands.
Moreover, according to SD-MVS \cite{SD-MVS}, SAM segmentation is proficient at extracting subtle edges while ignoring illumination disturbances compared with traditional edge detection.

Specifically, given the original image $I$ shown in Fig. \ref{fig: Diffusion} (a), we first adopt SAM2 for panoramic segmentation to obtain the segmentation image in Fig. \ref{fig: Diffusion} (b), denoted as $\mathcal{M}$. Subsequently, we apply a transformation function $\tau$ to extract the boundary of the segmentation image $\mathcal{M}$, thereby obtaining the boundary map $\mathcal{B}$, where $\mathcal{B} = \tau(\mathcal{M})$. 
Since depth edges in $\mathcal{B}$ are commonly located between distinct instances, we utilize $\mathcal{B}$ as the depth-edge guidance for subsequent patch deformation.

\subsubsection{Multi-Trajectory Diffusion Strategy}
As shown in Fig. \ref{fig: Diffusion} (g), SD-MVS \cite{SD-MVS} utilizes the depth-edge guidance by calculating the distance of each pixel to its boundaries only in both vertical and horizontal directions (up, down, left, and right), and then scales the corresponding patch based on the ratio of these distances. 
However, when faced with complicated scenarios which contain abundant instances with irregular shapes, such deformed patches typically struggle to accurately align with the boundaries of such irregular shapes, thereby leading to suboptimal deformations.

Therefore, how to ensure the deformation patch can comprehensively align with depth edges in complicated scenarios? 
Considering such a phenomenon: when a lamp is turned on in a dark room, each light emitted by its source will spread through the air and reach every opaque object’s surface without leaving any gaps. By analogy, if we respectively regard the light source, air, and object surface as the patch center, textureless areas, and depth edges, we can adopt a light diffusion-like approach to achieve edge-aligned patch deformation. 
However, practically enumerating the distances between the pixels and their edges in all directions for deformation is inappropriate, as it would be computationally inefficient. Therefore, we propose a multi-trajectory diffusion strategy for patch deformation, which not only achieves comprehensive edge alignment for patches but also achieves a compromise between computational complexity and effectiveness. 

Specifically, as shown in Fig. \ref{fig: Diffusion} (h), given the centered pixel $p$ in image $I$, we first emit $X$ rays around $p$ as a set of trajectories, notated as $\mathcal{L} = \{l_i | i = 1, ..., X \}$. Each trajectory $l_i$ is separated by an angle of ${\frac{360}{X}}^{\circ}$. $X$ is empirically set by 16 in our implementation. Then each trajectory $l_i$ will continue to extend until they intersect the boundary map $\mathcal{B}$ at boundary pixel $b_i$. Then we define the set of all boundary pixel $\mathcal{C}$ as: 
\begin{equation}
\mathcal{C} = \{b_i | b_i = l_i \cap \mathcal{B}, i = 1, ..., X \}.
\end{equation}
By connecting each intersection pixel $b_i$ in $\mathcal{C}$ we form the deformed patch for pixel $p$. Compared with SD-MVS in Fig. \ref{fig: Diffusion} (g), our method in Fig. \ref{fig: Diffusion} (h) can not only effectively enhance the reception field but also achieve an accurate boundary alignment for deformed patches, thereby allowing a flexible adaption to complicated scenarios containing the objects with irregular shapes. Built upon this strategy, we propose corresponding matching cost, propagation, and refinement strategies for further optimization steps, as detailed below.

\subsubsection{Patch Deformation on Matching Cost}
Although the multi-trajectory diffusion strategy enables deformed patches to align with edges in complicated scenarios, such deformation probably results in non-uniform patch areas. 
Therefore, if we apply traditional fixed-interval sampling for matching cost, it will inevitably cause time redundancy and imbalance.
However, according to API-MVS \cite{API-MVS}, we observe that by sampling only a few reliable pixels within the patch, we can approximately capture features of the entire patch.
Inspired by this, we propose a corresponding sampling scheme based on all trajectories in $\mathcal{L}$ to address the imbalance issue caused by fixed-interval sampling.

Specifically, as shown in Fig. \ref{fig: Diffusion} (i), we first calculate the average path length $|l_v|$ of all trajectories in $\mathcal{L}$. Then, we assign $n_i$ sampling pixels to each path $l_i$ for matching cost, where $n_i = \lceil \frac{l_i}{|l_v|} + \frac{1}{2} \rceil$.
Subsequently, to assign $n_i$ sampling pixels to each path $l_i$, the algorithm divides the path into $n_i$ fragments, each denoted as $S_{ij}$, where $j \in \{1, 2, \dots, n_i\}$. Then all pixel within each fragment $S_{ij}$ can be defined by:
\begin{equation}
S_{ij} = \{p_{ijk} \mid k = 1, 2, \dots, m_{ij}\},
\end{equation}
where $p_{ijk}$ represents the $k$-th pixel in the $j$-th fragment of the $i$-th trajectory, and $m_{ij}$ denotes the number of pixels in fragment $S_{ij}$.
To allocate these sampling pixels, the algorithm selects the pixel $p_{ij}^{\prime}$ with the minimal aggregated cost $c_a(p)$ from each segment $S_{ij}$, defined by:
\begin{equation}
p_{ij}^{\prime} = \arg\min_{p \in S_{ij}} c_a(p),
\label{1}
\end{equation}
where $c_a$ denotes the aggregated cost, whose detailed definition can be found in Eqn. \ref{9}.
Then the set $\mathcal{C}_i^{\prime}$ of all pixels selected for matching cost within the path $l_i$ is formulated by:
\begin{equation}
\mathcal{C}_i^{\prime} = \{p_{ij}^{\prime} \mid j = 1, 2, \dots, n_i\}, 
\end{equation}
where $n_i$ represents the number of sampling pixels assigned to path $l_i$.
Finally, by aggregating the collection $\mathcal{C}_i^{\prime}$ for each trajectory $l_i$ within $\mathcal{L}$ we can obtain the collection of all pixels selected for matching cost.
This sampling strategy guarantees that each path contributes at least one pixel. The longer trajectories will be assigned with more sampling pixels and vice versa, achieving a balanced and efficient allocation of sampling pixels for matching cost.


However, such sampling strategy often causes the failure of selecting enough reliable pixels for matching cost when well-textured pixels inside the deformed patch are not covered by the trajectory.
To address this issue, we propose a texture-aware mapping strategy to ensure the deformed patch acquires enough reliable pixels through mapping for robust matching cost.
Specifically, as shown in the white box of Fig. \ref{fig: Diffusion} (i), we first introduce pixel-wise textureness coefficient proposed in TAPA-MVS \cite{TAPA-MVS} as $t_p$. 
Then for each pixel $p \in S_{ij}$, we define its mapping pixel $m_p$ as the pixel with the minimum textureness coefficient $t_p$ within a $w \times w$ window $W_p$ centered at $p$, which is formulated as $m_p = \arg\min_{q \in W_p} t_q$.
Then the sampling pixel $p_{ij}^{\prime}$ in segment $S_{ij}$ is now determined by selecting the pixel with the minimal aggregated cost $c_a(p)$ among the mapping pixels of all pixels in $S_{ij}$, formulated as:
\begin{equation}
p_{ij}^{\prime\prime} = \arg\min_{p \in S_{ij}} c_a(m_p).
\label{2}
\end{equation}
Finally, by aggregating $p_{ij}^{\prime\prime}$, we obtain a more reliable sampling set $\mathcal{C}_i^{\prime\prime}$ for each trajectory $l_i$ for matching cost. Compared to $p_{ij}^{\prime}$ in Eqn. \ref{1}, $p_{ij}^{\prime\prime}$ in Eqn. \ref{2} significantly enhances the quality of patch features by extending the pixel sampling range. In summary, by combining both multi-trajectory diffusion and texture-aware mapping strategies, we not only achieve edge-aligned patch deformation but also select enough balanced and reliable sampling pixels for robust matching cost.

\subsubsection{Patch Deformation on Propagation and Refinement}
Concerning propagation, similar to SD-MVS \cite{SD-MVS}, we incorporate each trajectory and its diagonal trajectory to obtain eight combined trajectories. Then we select the pixel with the minimal aggregated cost $C_a$ on each combined trajectory for propagation, as shown in Fig. \ref{fig: Diffusion} (j),
Such propagation strategy not only ensures depth continuity within the searching space but also balances the search domain between different paths.

Concerning refinement, we adopt the spherical gradient refinement proposed in SD-MVS \cite{SD-MVS} for refinement, which adopts both spherical coordinates and gradient descent on normals, while employing pixel-wise searching intervals to constrain depths, respectively, thereby effectively enhancing the reliability for pixel-wise refined hypothesis. 

Obviously, the depth map adopted in our proposed deformation strategy shown in Fig. \ref{fig: Diffusion} (e) can effectively reconstruct textureless areas while preserving fine-grained details compared with the depth map of APD-MVS\cite{APD-MVS} which doesn't apply our proposed deformation strategy shown in Fig. \ref{fig: Diffusion} (d). This result demonstrates that the effectiveness of our proposed segmentation-driven and edge-aligned patch deformation.


\begin{figure*}
\centering
\includegraphics[width=\linewidth]{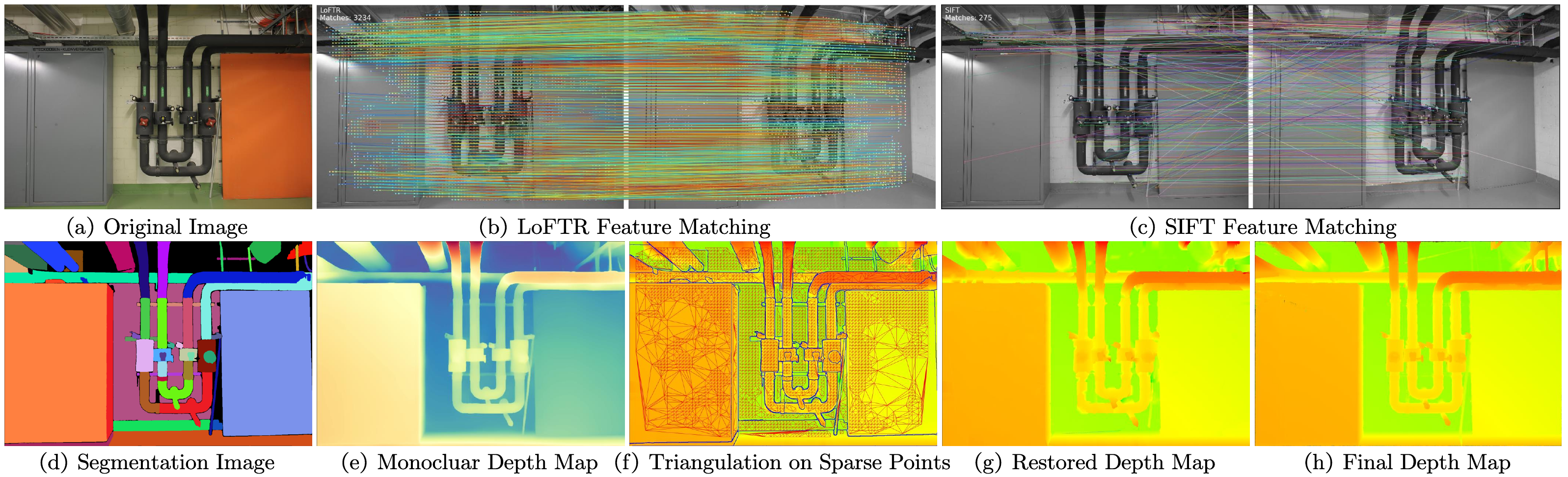}
\caption{
Sparse-monocular synergistic restoration. 
In (b) and (c), feature point pairs are linked with lines of different colors. 
In (d), different colors represent distinct instance masks.
In (e), lower color temperatures indicate greater depth values. 
In (f), instance edges and generated triangles are respectively highlighted in blue and red.
Compared with the final depth map in (h), the restored depth map in (g) effectively recovers textureless areas without severe detail distortion.
}
\label{fig: Triangulation}
\end{figure*}

\subsection{Sparse-Monocular Synergistic Restoration} 


Most PatchMatch algorithms achieve depth prediction by randomly initializing depth values within a predefined depth range, followed by propagation and refinement based on these initializations for depth estimation.
Consequently, a reliable initial depth map can significantly improve depth estimation accuracy.
Therefore, Mar-MVS \cite{MAR-MVS} adopts sparse points gained from COLMAP \cite{COLMAP} to constrain pixel-wise depth range for random initialization.
However, such random initialization heavily depends on the reliability of depth range, potentially causing the estimated depth to fall into a local minimum.

In contrast, Monocular Depth Estimation (MDE) can produce geometrically credible depth maps that are closely consistent with actual silhouettes of each instance. However, without inter-view projections, MDE can only yield relative depth rather than absolute depth.
Thus, applying relative-depth-based MDE directly to absolute-depth-based MVS for initialization may cause potential scale ambiguity. 
To eliminate this scale ambiguity, we propose a synergetic strategy to synthesize a high-confidence depth initialization. Firstly, we utilize a novel vision transformer-based feature matching foundation model called LoFTR \cite{loftr} to capture sufficient depth-reliable sparse points rapidly across diverse scenarios.  
Secondly, we propose to synergy both the depth-reliable sparse points from LoFTR and geometrically realistic depth maps from MDE for the depth map initialization. This will produce an optimal initialization for the further optimization steps.
For the detailed implementations of this strategy, we introduce the sparse-monocular synergistic restoration, which adopts segmentation-driven triangulation on sparse points and geometry-aware refinement on monocular depth maps to synergistically restore depth map for initialization and supervised guidance.





\subsubsection{Segmentation-driven Triangulation}
Specifically, as shown in Fig. \ref{fig: Triangulation} (b), we first employ LoFTR to extract features between reference image $I_i$ and source image $I_j$ to yield feature points $F_{ij}$.  
Compared to SIFT feature matching algorithm in Fig. \ref{fig: Triangulation} (c), LoFTR in Fig. \ref{fig: Triangulation} (b) offers superior speed and robustness, enabling the extraction of abundant feature points for restoration. 
Subsequently, following the pipeline of COLMAP \cite{COLMAP}, all feature points $F_{ij}$ associated with different image pairs between the reference image $I_i$ and the source image $I_j$ undergo triangulation, bundle adjustment, and sparse point cloud generation to produce all sparse points $P_i$ for reference image $I_i$. 
Next, segmentation-driven triangulation is applied to the sparse points $P_i$ to construct a coarse depth map. This map assumes the scene is composed of triangulated planes, thus its depth can be restored through triangulation on depth-reliable sparse points. 
Additionally, we further adopt SAM2 \cite{SAM2} to extract depth edges as guidance, preventing triangulation across instances with different depths.

Specifically, given the original image in Fig. \ref{fig: Triangulation} (a), we first adopt SAM2 for panoramic segmentation to obtain the segmentation image in Fig. \ref{fig: Triangulation} (d) and corresponding boundary map, which is then adopted to cluster the sparse points $P_i$ into groups $\{G_{ij}\}_{j=1}^J$, where $J$ represents the number of sparse point clusters.  
Triangulation is then performed within each cluster $G_{ij}$ to produce a series of triangle sets $\{T_{ijk}\}_{k=1}^K$, where $K$ indicates the number of triangles. Then for each pixel $p$ within the triangle $T_{ijk}$ whose vertices are respectively $T_A$, $T_B$ and $T_C$, its depth $d_p$ is formulated by:
\begin{equation}
d_p = \frac{\sum_{v \in \{T_A,T_B,T_C\}} d_v \cdot l_{vp}^{-1}}{\sum_{v \in \{T_A,T_B,T_C\}} l_{vp}^{-1}}\text{, if } p \in T_{ijk},
\end{equation}
where $d_v$ and $l_{vp}$ respectively represent the depth of vertex $v$ and the distance from vertex $v$ to pixel $p$. 
Moreover, for any pixel $p$ outside the triangulated sets $\{T_{ijk}\}_{k=1}^K$, a KD-tree \cite{KD-tree} is used to locate the nearest triangle $T_{ijk}$ within its same instance. Then its depth $d_p$ is determined by projecting pixel $p$ onto the plane of its nearest triangle $T_{ijk}$.
As shown in Fig. \ref{fig: Triangulation} (f), the proposed segmentation-driven triangulation effectively generates a coarse depth map that not only planarize textureless areas but also restores roughly accurate depth for other areas, building a solid foundation for further refinement.

\subsubsection{Geometry-aware Refinement}
Moreover, we further perform geometry-aware refinement on monocular depth map to provide the coarse depth map with geometry perception. Specifically, given the reference image in Fig. \ref{fig: Triangulation} (a), we first adopt DepthAnything V2 \cite{depth} to generate its monocular depth map $\mathcal{D}$ shown in Fig. \ref{fig: Triangulation} (e). DepthAnything V2 achieves higher accuracy compared to other MDE algorithms such as Marigold \cite{Marigold} and ZoeDepth \cite{ZoeDepth}, making it more suitable for complicated scenarios. 
Subsequently, to validate the reliability of each identified triangle $T_{ijk}$, we perform RANSAC-based planar fitting on its corresponding pixels in the monocular depth map $\mathcal{D}$ based on threshold $\gamma$, thus obtaining an inlier ratio $r_k$ and estimated plane $\pi_k$. 
Then if $r_k$ exceeds threshold $\kappa$, the triangle $T_{ijk}$ is classified as a planar area, with its depth remains unchanged. Otherwise, the triangle $T_{ijk}$ is considered as non-planar area with geometric complexity, with its depth adjusted based on the relative depth differences in monocular depth map $\mathcal{D}$. Specifically, for any pixel $p$ within non-planar triangle $T_{ijk}$ whose vertices are respectively $T_A$, $T_B$ and $T_C$, its depth $d_p$ is computed as:
\begin{equation}
d_p = \frac{\sum_{v \in \{T_A,T_B,T_C\}} d_v \cdot m_{vp}^{-1}}{\sum_{v \in \{T_A,T_B,T_C\}} m_{vp}^{-1}}\text{, if } p \in T_{ijk},
\label{7}
\end{equation}
where $d_v$ and $m_{vp}$ respectively represent the depth of vertex $v$ and the depth different from vertex $v$ to pixel $p$ in the monocular depth map $\mathcal{D}$. Similarly, for any pixel $p$ outside the triangulated sets $\{T_{ijk}\}_{k=1}^K$, we adopt the KD-tree \cite{KD-tree} to locate its nearest planar triangle $T_{ijk}$ within a same instance. 
Then the algorithm evaluates the distance between pixel $p$ and the plane $\pi_k$ of its nearest planar triangle $T_{ijk}$. 
If this distance is less than $\kappa$, $p$ is considered to belong to textureless areas, with its depth remaining unchanged.

Otherwise, we consider pixel $p$ located in non-planar area, with its depth $d_p$ adjusted with a proportional mapping algorithm. Specifically, we first identity the nearest pixel $q$ to $p$ within the triangle $T_{ijk}$, then the depth $d_p$ of pixel $p$ is defined as: $d_p=d_q \cdot D_p / D_q$, where $d_q$ is the depth of pixel $q$ obtained through Eqn. \ref{7}, $D_p$ and $D_q$ respectively represent the depths of $p$ and $q$ in the monocular depth map $\mathcal{D}$. Through such a simply proportional mapping equation, we subtly leverage the depth variation in the monocular depth map $\mathcal{D}$ to estimate the depth in non-planar areas with geometric complexity.

Compared to the final depth map generated by our method shown in Fig. \ref{fig: Triangulation} (h), our restored depth map shown in Fig. \ref{fig: Triangulation} (g) is both reliable and realistic, effectively recovering the majority of both textureless and well-textured areas. 

\subsubsection{Initialization and supervision}
After obtaining the restored depth map, we respectively adopt it for both initialization and supervised guidance during the patch match (PM) process. 
Specifically, considering that a reliable initial depth map can effectively aid in reconstruction, we first adopt the restored depth map to replace the original randomly initialized depth map for initialization, thus effectively improving the accuracy for subsequent propagation and refinement.

Moreover, since other methods lack reliable pixel-wise depth constraint, their estimated depths may diverge significantly from the actual depth due to getting stuck in local minima. 
Addressing this, the algorithm incorporates the restored depth map into the EM-based optimization algorithm as supervised guidance, thereby ensuring the depth range remains valid during both propagation and refinement.

Specifically, during the patch-matching phase, we consider not only the multi-scale matching cost $C_{m}$, but also the reprojection error $E_{re}$ and the projection color gradient error $E_{cl}$.
The multi-scale matching cost $C_{m}$ is computed by uniformly aggregating the matching costs across all layers.
The reprojection error $E_{re}$ is utilized to validate the robustness of depth estimation from geometric consistency.
It first projects pixel $p_i$ with its estimated depth from the reference image $I_i$ into the source image $I_j$ to obtain its projection pixel $p_j$. Since $p_j$ has a corresponding estimated depth in $I_j$, we adopt it to reproject $p_j$ back into $I_i$, thereby deriving reprojection pixel $p_i^{\prime}$. $E_{re}$ is then derived by $L_2$-norm formulated as follows:
\begin{equation}
E_{re}=\min \left(\left\|p_i^{\prime}-p_i\right\|_2, \tau\right), 
\end{equation}
where $\tau$ is regularization parameter to suppress the outliers.
Moreover, the projection color gradient error $E_{cl}$ measures color consistency between current pixel $p_i$ in reference image $I_i$ and its projection pixel $p_j$ in source images $I_j$:
\begin{equation}
    E_{cl} = \max \left( \| \nabla I_i\left( p_i \right) - \nabla I_j\left( p_j \right) \|_2, \tau \right),
\end{equation}
where $\nabla$ represents the Laplacian Operator, $p_j$ denotes pixel in image $I_j$ the projected by pixel $p_i$ in $I_i$. 
Additionally, to prevent depth estimation from falling into local minima, we further incorporate the restored depth map as supervised guidance. Specifically, a depth difference error $E_{dp}$ is introduced to measure the reliability of depth $d_p$ for pixel $p$, defined by:
\begin{equation}
E_{dp}= \begin{cases}0 & \text { if } \frac{\left|d_p-d_p^{\prime}\right|}{d_p} \leq \mu, \\ 
1 & \text { otherwise, }
\end{cases}
\end{equation}
where $d_p$ represents the depth of pixel $p$ in the restored depth map, and $\mu$ is the truncation threshold. We define $\mu = 0.05 \times 2^{n}$ where $n$ denotes the current PM layer, such that the algorithm allows for greater depth tolerance as iteration progresses.
Finally, by aggregating all the above-mentioned terms, the aggregated costs $c_{a}$ can be given by:
\begin{equation}
    C_{a}=w_{m}C_{m}+w_{r}E_{re}+w_{c}E_{cl}+w_{d}E_{dp},
\label{9}
\end{equation}
where $w_{m}$, $w_{r}$, $w_{c}$, and $w_{d}$ respectively represent the aggregation weights of each component. 
Subsequently, similar to SD-MVS \cite{SD-MVS}, during the propagation and refinement stages, the algorithm fixes 
$w_{m}$, $w_{r}$, $w_{c}$, and $w_{d}$ to minimize the aggregated costs $c_{a}$ as E-step, formulated as:
\begin{equation}
\underset{C_{m},E_{re},E_{cl},E_{dp}}{\min} C_{a}=w_{m}C_{m}+w_{r}E_{re}+w_{c}E_{cl}+w_{d}E_{dp}.
\end{equation}
Then between each PM iteration, the algorithm fixes $c_{m}$, $e_{r}$, $e_{p}$, and $e_{g}$ to optimize each parameter as M-step, defined as: 
\begin{equation}
\begin{aligned}
\min _{w_{m},w_{r},w_{c},w_{d}} & C_{a}=w_{m}C_{m}+w_{r}E_{re}+w_{c}E_{cl}+w_{d}E_{dp}, \\
\text { s.t. } & w_{m}+w_{r}+w_{c}+w_{d}=1, \\
& w_{m},w_{r},w_{c},w_{d} > \eta.
\end{aligned}
\end{equation}
All hyperparameters are required to exceed a minimal value $\eta$, and we implement a normalization constraint ensuring that their sum equals $1$ to mitigate significant variances.
Following the E-step optimization, we can alternatively optimize the hyperparameters and feed them back into the E-step for the next round of aggregated cost optimization.
Therefore, by doing so, we can effectively leverage pixel-level reliable depth as guidance, thus preventing erroneous estimations caused by depths being trapped in local minima.

\begin{figure*}
\centering
\includegraphics[width=\linewidth]{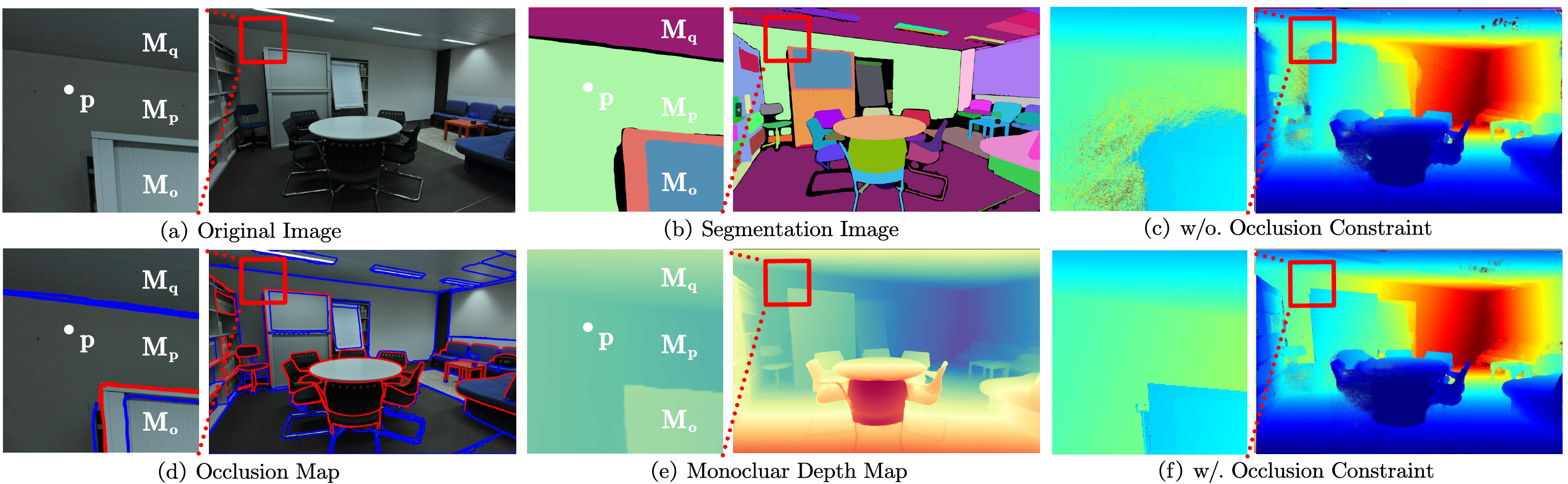}
\caption{
Occlusion-aware patch deformation. In (b), different colors represent different instance masks. 
In (c) and (f), higher color temperatures indicate greater depth values. 
Differently, in (e), lower color temperatures indicate greater depth values. 
In (d), blue and red edges respectively denote depth-continuous and depth-discontinuous boundaries. 
Compared with (c), depth map with occlusion constraint in (f) can effectively distinguish depth edges in textureless areas.
}
\label{fig: Occlusion}
\end{figure*}

\subsection{Occlusion-Aware Patch Deformation}
Similar to the previous sections, we first adopt SAM2 on the original image for panoramic segmentation to extract depth edges as guidance for patch deformation.
For instance, as shown in Fig. \ref{fig: Occlusion} (a), during patch deformation, pixel $p$ in instance $M_p$ is limited to only referencing its surrounding areas without crossing depth edge boundaries.
However, when its surrounding areas are entirely composed of textureless areas without well-textured pixels, pixel $p$ inevitably struggles to identify enough reliable pixels for reconstruction, potentially leading to matching ambiguity. Consequently, strictly applying SAM-based edges as a constraint for patch deformation may not always yield optimal results.

Therefore, how can we distinguish such cases? As shown in Fig. \ref{fig: Occlusion} (a), among the two instances $M_q$ and $M_o$ adjacent to the instance $M_p$ of pixel $p$, instance $M_q$ does not occlude $M_p$, thus its depth remains continuous near their shared boundary. Therefore, its reliable pixels near the boundary can theoretically be selected for patch deformation of pixel $p$ in $M_p$. Conversely, since instance $M_o$ occludes $A$, there is a depth discontinuity across their boundary, making its reliable pixels near the boundary unsuitable for guiding the patch deformation of pixel $p$ in $M_p$.

From the analysis, we understand that by identifying the occlusion relationship between different instances, the algorithm can selectively assign edge constraints with different intensities for patch deformation. Such an idea effectively addresses the limitations of suboptimal deformation under previous rigid edge constraint strategy.
Therefore, we integrate the segmentation map with the monocular depth map to exploit inter-instance occlusion relationship, then regard them as occlusion map to implement two distinct edge constraints, thus facilitating occlusion-aware patch deformation.

\subsubsection{Occlusion Map Generation}
Specifically, according to the preceding content, we can utilize the original image in Fig. \ref{fig: Occlusion} (a) to generate corresponding segmentation image $\mathcal{M}$ in Fig. \ref{fig: Occlusion} (b), the boundary map $\mathcal{B}$ and the monocular depth map $\mathcal{D}$ in Fig. \ref{fig: Occlusion} (e). 
Then for each boundary pixel $b$ within the boundary map $\mathcal{B}$, we construct a window $W_b$ centered at $b$ with size $w \times w$. 
Subsequently, for each pixel within the window $W_b$, we adopt the Sobel operator to compute its depth gradient in the monocular depth map $\mathcal{D}$ and select the maximum gradient value $G_{\max}$ for identification, defined as:
\begin{equation}
G_{\max }=\max _{p \in W_b} \sqrt{\left(\frac{\partial D(p)}{\partial x}\right)^2+\left(\frac{\partial D(p)}{\partial y}\right)^2}.
\end{equation}
Then we define the label of boundary pixel $b$ as depth-continuous and mark it as a blue dot when $G_{\max} < \delta$. Otherwise, we consider it as depth-discontinuous and mark it as a red dot, thereby generating the occlusion map shown in Fig. \ref{fig: Occlusion} (d).
Note that to address the misclassification of depth continuity caused by noise disturbance, we connect all boundary pixels with the same color into different clusters and then reassign clusters whose number of boundary pixels is fewer than the threshold $\sigma$ to the opposite color. 
Such a strategy can effectively correct the misclassification of partial noise-disturbed boundary pixels.
Consequently, the final occlusion map in Fig. \ref{fig: Occlusion} (d) can effectively distinguish the depth continuity across boundaries, thereby exploring the occlusion relationships between different instances for the subsequent edge constraint strategy.

\subsubsection{Dual-Categories Edge Constraint}
After obtaining the occlusion map, we further propose a dual-categories edge constraint to enable patch deformation with occlusion perception. 
Specifically, as shown in Fig. \ref{fig: Occlusion} (d), during the patch deformation process of pixel $p$, if it encounters red edges, since red edges represent depth discontinuities, the algorithm assumes a significant occlusion relationship between adjacent instances. 
Consequently, a strict edge constraint is applied, ensuring that both the path $l_i$ used for propagation, refinement and the mapping pixels $m_p$ used for matching cost are restricted from crossing the edge.

In contrast, when encountered edges are blue, as blue edges represent depth continuity, the algorithm assumes no distinct occlusion between adjacent instances. 
Therefore, a flexible edge constraint is applied, allowing both the path $l_i$ used for propagation, refinement and the mapping pixels $m_p$ used for matching cost to cross the edge within a specified threshold $\varepsilon$. 
This strategy ensures reliable pixels from other instances can also be considered during matching cost, propagation and refinement, thus effectively enhancing the perception of deformed patched and the efficiency of both propagation and refinement.
By doing so, the algorithm not only explores inter-instance occlusion relationship but also addresses the potential deformation instability caused by prior edge constraints, thereby achieving occlusion-aware patch deformation.

\section{Experiments}
We evaluate our method on four different datasets, the ETH3D high-resolution benchmark \cite{ETH3D}, the Tanks \& Temples benchmark (TNT) \cite{TNT}, the BlendedMVS dataset \cite{Blendedmvs} and the Strecha dataset \cite{strecha}. Specifically, we first conduct both qualitative and quantitative results of our point clouds against other methods among the above-mentioned datasets for evaluation. 
Then we perform ablation studies on each proposed module and most parameters to demonstrate the effectiveness of module and justify our parameter choices. 
Moreover, we conduct a comparative analysis of GPU memory usage and runtime to evaluate the efficiency of our method. 
Extensive evaluations indicate that our method can achieve state-of-the-art performance across both traditional and learning-based approaches with strong generalization capabilities. 


\begin{table*}
  \centering
  \renewcommand{\arraystretch}{1.05} 
      \captionsetup{labelfont={color=black}}
      \caption{Quantitative results on ETH3D dataset at threshold $2cm$ and $10cm$.}
    \resizebox{0.9\linewidth}{!}{
        \begin{tabular}{c|ccc|ccc|ccc|ccc}
        \hline
        \multirow{3}{*}{Method} & \multicolumn{6}{c|}{Train} & \multicolumn{6}{c}{Test} \\
        \cline{2-13} & \multicolumn{3}{c|}{$2cm$} & \multicolumn{3}{c|}{$10cm$} & \multicolumn{3}{c|}{$2cm$} & \multicolumn{3}{c}{$10cm$}  \\
        \cline{2-13} & F$_1$ & Comp. & Acc. & F$_1$ & Comp. & Acc. & F$_1$ & Comp. & Acc. & F$_1$ & Comp. & Acc. \\   
        \hline 
        \multirow{1}{*}{PatchMatchNet\cite{Iter-MVS}} & 64.21 & 65.43 & 64.81 & 85.70 & 83.28 & 89.98 & 73.12 & 77.46 & 69.71 & 91.91 & 92.05 & 91.98 \\   
        \multirow{1}{*}{IterMVS-LS\cite{Iter-MVS}} & 71.69 & 66.08 & 79.79 & 88.60 & 82.62 & 96.35 & 80.06 & 76.49 & 84.73 & 92.29 & 88.34 & 96.92 \\   
        \multirow{1}{*}{MVSTER\cite{MG-MVS}} & 72.06 & 76.92 & 68.08 & 91.73 & 91.91 & 91.97 & 79.01 & 82.47 & 77.09 & 93.20 & 92.71 & 94.21 \\  
        \multirow{1}{*}{EPP-MVSNet\cite{EPP-MVSNet}} & 74.00 & 67.58 & 82.76 & 92.13 & 87.72 & 97.29 & 83.40 & 81.79 & 85.47 & 95.22 & 93.75 & 96.84 \\  
        \multirow{1}{*}{EP-Net\cite{EP-Net}} & 79.08 & 79.28 & 79.36 & 93.92 & 93.69 & 94.33 & 83.72 & 87.84 & 80.37 & 95.20 & 96.82 & 93.72 \\         
        \hline 
        \multirow{1}{*}{TAPA-MVS\cite{TAPA-MVS}} & 77.69 & 71.45 & 85.88 & 93.69 & 90.98 & 96.79 & 79.15 & 74.94 & 85.71 & 92.30 & 90.35 & 94.93 \\
        \multirow{1}{*}{PCF-MVS\cite{PCF-MVS}} & 79.42 & 75.73 & 84.11 & 92.98 & 90.42 & 95.98 & 80.38 & 79.29 & 82.15 & 91.56 & 91.26 & 92.12 \\
        \multirow{1}{*}{ACMMP\cite{ACMMP}} & 83.42 & 77.61 & \textcolor{red}{\textbf{90.63}} & 95.54 & 93.32 & \textbf{97.99} & 85.89 & 81.49 & \textcolor{red}{\textbf{91.91}} & 96.27 & 94.67 & \textcolor{red}{\textbf{98.05}} \\   
        \multirow{1}{*}{APD-MVS\cite{APD-MVS}} & 86.84 & 84.83 & 89.14 & 97.12 & 96.79 & 97.47 & 87.44 & 85.93 & 89.54 & 96.95 & 96.95 & 97.00 \\   
        \multirow{1}{*}{HPM-MVS++\cite{HPM-MVS}} & \textbf{87.09} & \textbf{85.64} & 88.74 & 97.22 & \textbf{96.91} & 97.56 & \textbf{89.02} & \textbf{89.37} & 88.93 & 97.34 & \textbf{97.72} & 96.99 \\  
        \multirow{1}{*}{SD-MVS\cite{SD-MVS} (base)} & 86.94 & 84.52 & 89.63 & \textbf{97.35} & 96.87 & 97.84 & 88.06 & 87.49 & 88.96 & \textbf{97.41} & 97.51 & 97.37 \\  
        \hline
        \multirow{1}{*}{SED-MVS (ours)} & \textcolor{red}{\textbf{88.85}} & \textcolor{red}{\textbf{88.94}} & \textbf{88.96} & \textcolor{red}{\textbf{97.98}} & \textcolor{red}{\textbf{97.92}} & \textcolor{red}{\textbf{98.05}} & \textcolor{red}{\textbf{90.08}} & \textcolor{red}{\textbf{89.94}} & \textbf{90.46} & \textcolor{red}{\textbf{98.05}} & \textcolor{red}{\textbf{98.18}} & \textbf{97.94} \\  
        \hline
        \end{tabular}%
    }
    \label{table:ETH3D}%
    \vspace{-0.09in}
\end{table*}%

\begin{table*}
  \centering
  \renewcommand{\arraystretch}{1.05} 
      \captionsetup{labelfont={color=black}}
  \caption{Quantitative results on partial scenes of Strecha dataset (\emph{Fountain} and \emph{HerzJesu}) at threshold $2cm$ and $10cm$.}
    \resizebox{0.9\linewidth}{!}{
        \begin{tabular}{c|ccc|ccc|ccc|ccc}
        \hline
        \multirow{3}{*}{Method} & \multicolumn{6}{c|}{Fountain} & \multicolumn{6}{c}{HerzJesu} \\
        \cline{2-13} & \multicolumn{3}{c|}{$2cm$} & \multicolumn{3}{c|}{$10cm$} & \multicolumn{3}{c|}{$2cm$} & \multicolumn{3}{c}{$10cm$}  \\
        \cline{2-13} & F$_1$ & Comp. & Acc. & F$_1$ & Comp. & Acc. & F$_1$ & Comp. & Acc. & F$_1$ & Comp. & Acc. \\   
        \hline 
        \multirow{1}{*}{OpenMVS\cite{OpenMVS}} & 74.77 & 70.47 & 79.62 & 87.37 & 83.49 & 91.63 & 69.67 & 61.85 & 79.76 & 79.94 & 70.37 & 92.53 \\
        \multirow{1}{*}{PatchMatchNet\cite{Iter-MVS}} & 69.06 & 68.73 & 69.4 & 84.61 & 82.13 & 87.25 & 62.43 & 59.32 & 65.89 & 77.11 & 69.25 & 86.98 \\  
        \multirow{1}{*}{IterMVS-LS\cite{Iter-MVS}} & 75.63 & 69.45 & 83.02 & 87.78 & 82.61 & 93.65 & 69.73 & 60.14 & 82.97 & 80.33 & 69.83 & 94.55 \\
        \multirow{1}{*}{MVSTER\cite{Iter-MVS}} & 76.05 & 78.62 & 73.65 & 89.72 & 90.85 & 88.61 & 71.04 & 71.43 & 70.65 & 83.26 & 79.04 & 87.95 \\   
        \hline 
        \multirow{1}{*}{ACMM\cite{ACMM}} & 75.48 & 67.32 & 85.89 & 87.28 & 81.29 & 94.23 & 68.24 & 57.56 & 83.78 & 79.11 & 67.85 & 94.86 \\
        \multirow{1}{*}{ACMP\cite{ACMP}} & 76.38 & 68.53 & 86.26 & 87.91 & 82.15 & 94.55 & 69.94 & 59.73 & 84.35 & 80.49 & 69.72 & 95.19 \\
        \multirow{1}{*}{ACMMP\cite{ACMMP}} & 80.47 & 73.75 & \textcolor{red}{\textbf{88.53}} & 89.88 & 84.36 & \textcolor{red}{\textbf{96.18}} & 73.13 & 63.41 & \textcolor{red}{\textbf{86.38}} & 83.01 & 72.65 & \textcolor{red}{\textbf{96.81}} \\
        \multirow{1}{*}{APD-MVS\cite{APD-MVS}} & 83.71 & 80.63 & 87.04 & 91.60 & 87.89 & 95.64 & 77.89 & 72.13 & 84.65 & 85.63 & 77.18 & 96.16 \\  
        \multirow{1}{*}{HPM-MVS++\cite{HPM-MVS}} & \textbf{84.02} & \textbf{81.84} & 86.32 & \textbf{91.73} & \textbf{88.03} & 95.75 & \textbf{78.41} & \textbf{73.69} & 83.77 & \textbf{85.78} & \textbf{77.34} & 96.28 \\  
        \multirow{1}{*}{SD-MVS\cite{SD-MVS} (base)} & 83.78 & 80.32 & \textbf{87.56} & 91.70 & 87.92 & 95.85 & 77.96 & 71.84 & \textbf{85.23} & 85.82 & 77.26 & 96.51 \\  
        \hline 
        \multirow{1}{*}{SED-MVS (ours)} & \textcolor{red}{\textbf{85.73}} & \textcolor{red}{\textbf{84.67}} & 86.82 & \textcolor{red}{\textbf{92.42}} & \textcolor{red}{\textbf{89.01}} & \textbf{96.11} & \textcolor{red}{\textbf{80.54}} & \textcolor{red}{\textbf{77.04}} & 84.37 & \textcolor{red}{\textbf{86.65}} & \textcolor{red}{\textbf{78.48}} & \textbf{96.72} \\  
        \hline 
        \end{tabular}%
    }
  \label{table:strecha}%
  \vspace{-0.09in}
\end{table*}%

\subsection{Datasets and Implementation Details}
The ETH3D high-resolution benchmark \cite{ETH3D} comprises 25 scenes, each containing images with a resolution of 6,221 × 4,146. The dataset presents significant challenges for reconstruction due to its varying discrete viewpoints and diverse scene types. It is split into both training and testing datasets. The training set, consisting of 13 scenes, includes both ground truth point clouds and depth maps, while the testing set, with 12 scenes, has its ground truth point clouds and depth maps retained by its benchmark’s official platform.

The Tanks and Temples (TNT) Benchmark \cite{TNT} comprises 14 unique scenes with a resolution of 1,920 × 1,080, covering a range of individual objects like francis, panther and large-scale indoor scenes like museum and ballroom. Note that its ground truth depth maps are captured through high-quality industrial laser scanners. Both the ETH3D and TNT benchmarks are divided into training and testing sets. We have uploaded our results to their official websites for public reference.

The BlendedMVS high-resolution dataset \cite{Blendedmvs} contains 17,000 multi-view stereo (MVS) training samples across 113 diverse scenes, including architectural structures, sculptures, and small objects, with a resolution of 2,048 × 1,536. Compared to other datasets, networks trained on this dataset exhibit superior generalization. We conduct qualitative experiments on this dataset to validate the robustness of our algorithm.

\begin{table}
  \centering
  \renewcommand{\arraystretch}{1.07} 
    \captionsetup{labelfont={color=black}}
    \caption{Quantitative results on TNT dataset at given threshold.}
    \resizebox{\linewidth}{!}{
        \begin{tabular}{c|ccc|ccc}
        \hline
        \multirow{2}{*}{Method} & \multicolumn{3}{c|}{Intermediate} & \multicolumn{3}{c}{Advanced} \\
        \cline{2-7} & F$_1$ & Rec. & Pre. & F$_1$ & Rec. & Pre. \\   
        \hline 
        \multirow{1}{*}{PatchMatchNet\cite{PatchMatchNet}} & 53.15 & 69.37 & 43.64 & 32.31 & 41.66 & 27.27 \\ 
        \multirow{1}{*}{IterMVS-LS\cite{Iter-MVS}} & 56.94 & 74.69 & 47.53 & 34.17 & 44.19 & 28.70 \\  
        \multirow{1}{*}{AGG-CVCNet\cite{AGG-CVCNet}} & 57.81 & 71.71 & 49.04 & 28.96 & 28.28 & 35.33 \\  
        \multirow{1}{*}{MVSTER\cite{MVSTER}} & 60.92 & \textcolor{red}{\textbf{77.50}} & 50.17 & 37.53 & 45.90 & 33.23 \\  
        \multirow{1}{*}{EPP-MVSNet\cite{EPP-MVSNet}} & 61.68 & 75.58 & 53.09 & 35.72 & 34.63 & \textbf{40.09} \\
        \multirow{1}{*}{EP-Net\cite{EP-Net}} & \textbf{63.68} & 72.57 & \textcolor{red}{\textbf{57.01}} & \textcolor{red}{\textbf{40.52}} & \textbf{50.54} & 34.26 \\ 
        \hline 
        \multirow{1}{*}{PCF-MVS\cite{PCF-MVS}} & 53.39 & 58.85 & 50.04 & 34.59 & 34.35 & 35.84 \\  
        \multirow{1}{*}{ACMP\cite{ACMP}} & 58.41 & 73.58 & 49.06 & 37.44 & 42.48 & 34.57 \\  
        \multirow{1}{*}{ACMMP\cite{ACMMP}} & 59.38 & 68.50 & 53.28 & 37.84 & 44.64 & 33.79 \\  
        \multirow{1}{*}{APD-MVS\cite{APD-MVS}} & 63.64 & 75.06 & 55.58 & 39.91 & 49.41 & 33.77 \\  
        \multirow{1}{*}{HPM-MVS++\cite{HPM-MVS}} & 61.59 & 73.79 & 54.01 & 39.65 & 41.09 & \textcolor{red}{\textbf{40.79}} \\  
        \multirow{1}{*}{SD-MVS\cite{SD-MVS} (base)} & 63.31 & 76.63 & 53.78 & 40.18 & 50.31 & 33.81 \\  
        \hline 
        \multirow{1}{*}{SED-MVS (ours)} & \textcolor{red}{\textbf{64.86}} & \textbf{77.09} & \textbf{56.20} & \textbf{40.41} & \textcolor{red}{\textbf{54.20}} & 32.37 \\  
        \hline
        \end{tabular}%
    }
  \label{table:TNT}%
  \vspace{-0.09in}
\end{table}%

The Strecha dataset \cite{strecha} includes six outdoor scenes with a resolution of 3,072 × 2,048. Among them, only the Fountain and HerzJesu scenes provide ground truth point clouds for evaluation, consisting of 11 and 8 images, respectively. 

\begin{figure*}
\centering
\includegraphics[width=\linewidth]{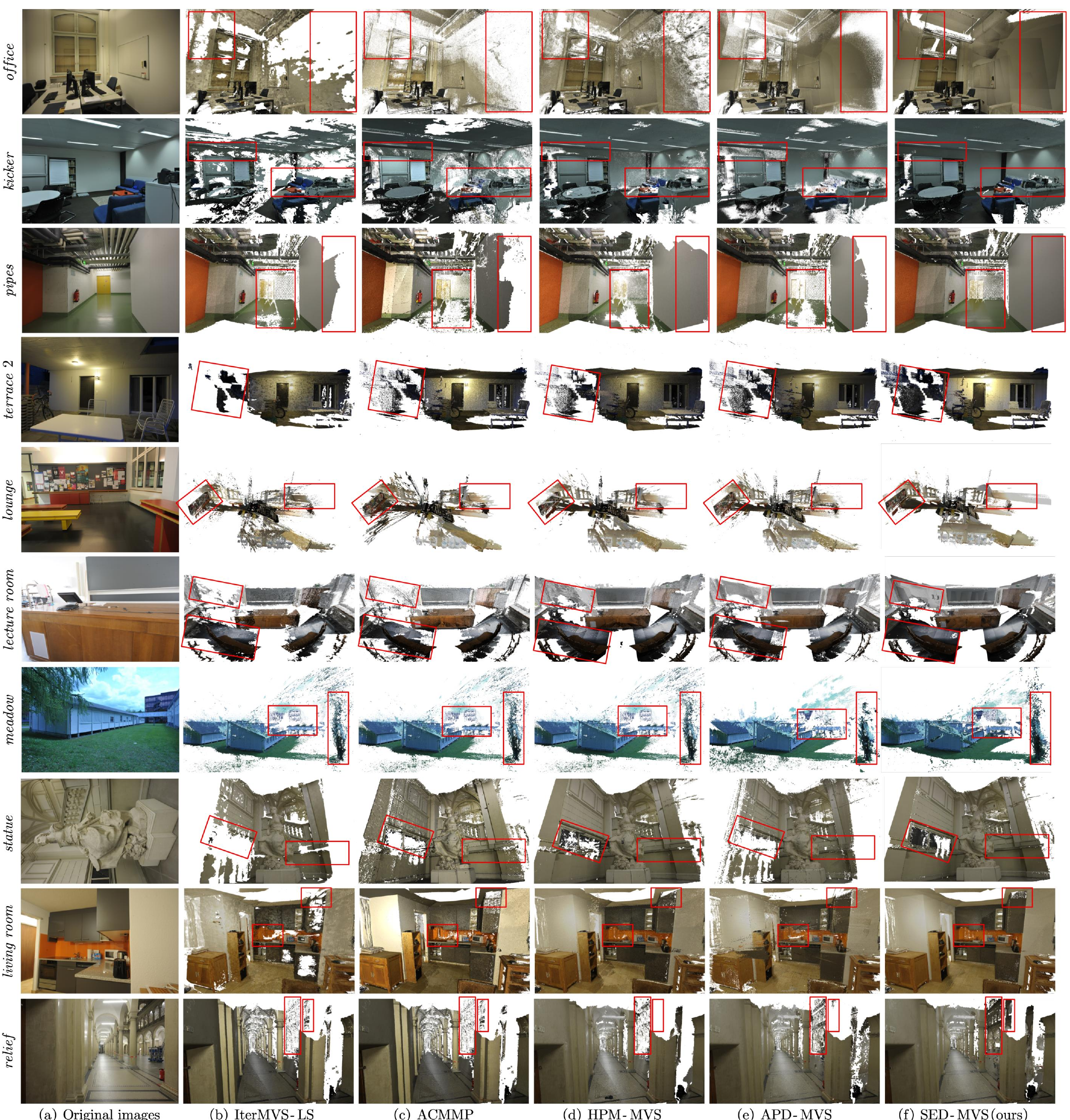}
\caption{
Visualized point cloud results between different methods on partial scenes of ETH3D datasets (\emph{office}, \emph{kicker}, \emph{pipes}, \emph{terrace 2}, \emph{lounge}, \emph{lecture room}, \emph{meadow}, \emph{statue}, \emph{living room} and \emph{relief}). Obviously, our method can effectively reconstruct textureless areas such as floors and walls without detail distortion. 
}
\label{fig: ETH3D}
\end{figure*}

\begin{figure*}
\centering
\includegraphics[width=\linewidth]{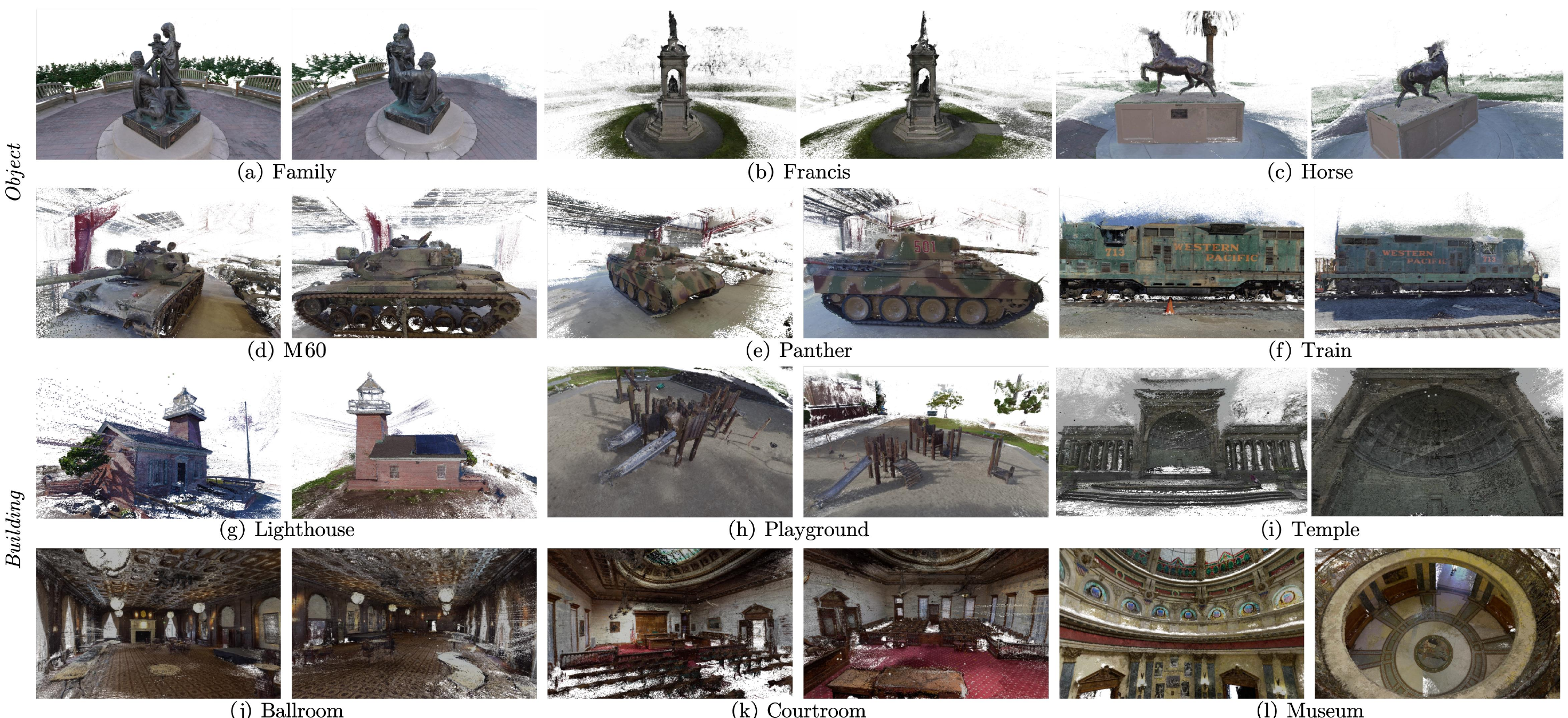}
\caption{
Reconstructed point clouds of our method on Tanks \& Temples dataset without any fine-tuning.
}
\label{fig: TNT}
\end{figure*}

In order to achieve faster times during actual experiments, we downsampled the images to half of the original resolution in both the ETH3D and Strecha datasets, whereas adopting the original resolution in both the TNT and BlendedMVS datasets. This adaptation does not compromise the robustness of our approach while enhancing its efficiency on runtime.

Our method is implemented on a system equipped with an Intel(R) Core(TM) i7-10700 CPU @ 2.90GHz and an NVIDIA GeForce RTX 3080 graphics card. Concerning the parameter selection, $\{X, w, \gamma, \kappa, \tau, \mu, \eta, \delta, \sigma, \varepsilon\} = \{16, 11, 5\times10^{-3}, 0.7, 3, 0.05\times2^{n}, 0.1, 1.8, 12, 8\times2^{n}\}$, where $n$ denote the current PM layer. Concerning EM optimization, the initial weights are $\{w_m, w_r, w_c, w_d\} = \{1, 0.2, 0.2, 0.2\}$.

\subsection{Point Cloud Evaluation}

To evaluate the effectiveness of our method, we compared the accuracy (Acc.), completeness (Comp.), and F1 score of the reconstructed point clouds with other approaches. 
we compare our method with traditional MVS algorithm such as ACMM \cite{ACMM}, ACMP \cite{ACMP}, ACMMP \cite{ACMMP}, and APD-MVS \cite{APD-MVS}, HPM-MVS++ \cite{HPM-MVS}, SD-MVS \cite{SD-MVS} as well as well-known learning-based MVS methods like PatchMatchNet \cite{PatchMatchNet}, IterMVS-LS \cite{Iter-MVS}, MVSTER \cite{MVSTER}, EPP-MVSNet \cite{EPP-MVSNet} and EP-Net \cite{EP-Net}.
Quantitative results for the ETH3D, Strecha, and TNT datasets are presented in Tab. \ref{table:ETH3D}, Tab. \ref{table:strecha} and Tab. \ref{table:TNT}, respectively. The best results are highlighted in bold and red, while the second-best results are marked in bold and black. 
For the ETH3D and TNT datasets, quantitative results of other methods are directly obtained from their official websites. Differently, for the Strecha datasets, we replicate the results of other methods for quantitative evaluation.

As shown in the above tables, for the ETH3D and Strecha datasets, our method achieves the highest F1 scores and completeness across both training and testing sets, validating its superior effectiveness. 
For the TNT dataset, we test our method without fine-tuning to demonstrate its generalization capability. On the intermediate set, our method achieved the highest F1 score and completeness. On the advanced set, our method delivered the highest completeness, only falling short by \textbf{0.11\%} in F1 score compared to EPNet, demonstrating the astonishing robustness of our proposed method.

Qualitative results for the ETH3D and TNT datasets are respectively shown in Fig. \ref{fig: ETH3D} and \ref{fig: TNT}.
As shown in Fig. \ref{fig: ETH3D}, our method produces the most complete and realistic reconstructed point clouds compared to other methods, particularly in textureless areas such as floors and walls (highlighted in red boxes), without introducing conspicuous detail distortion. Furthermore, as evident in Fig. \ref{fig: TNT}, our method can effectively reconstruct textureless areas while preserving fine-grained details, validating its robustness and generalization ability. More qualitative results on both the Strecha and BlendedMVS datasets along with a detailed memory and runtime comparison are shown in supplementary materials.

\subsection{Ablation Study}
To validate the devised modules of our method, we conduct a series of experiments for ablations study to illustrate the effectiveness of each proposed module, as shown in Tab. \ref{table: ablation study}. 
\subsubsection{Segmentation-Driven and Edge-Aligned Deformation} 
We separately exclude the whole segmentation-driven and edge-aligned patch deformation module (w/o. Def.), multi-trajectory diffusion strategy (w/o. Taj.), texture-aware mapping (w/o. Map.), spherical gradient refinement (w/o. Ref.) and load balancing propagation (w/o. Pro.) for ablations. Notably, w/o. Def. yields the worst F$_1$ score, validating the significance of segmentation-driven and edge-aligned patch deformation strategy. The F$_1$ score of w/o. Taj. is better than that of w/o. Map., emphasizing that multi-trajectory diffusion strategy plays a more critical role than texture-aware mapping, though both make sufficiently essential contributions. 
Although both w/o. Ref. and w/o. Pro. achieve similar F$_1$ score, w/o. Ref. exhibits lower accuracy but higher completeness, suggesting that load balancing propagation demonstrates superiorities in reconstructing textureless areas, whereas spherical gradient refinement can effectively improve reconstruction precision.

\begin{table}
    \centering
    \renewcommand{\arraystretch}{1.07} 
    \caption{Quantitative results of the ablation studies on ETH3D benchmark to validate each proposed component.}
    \resizebox{\linewidth}{!}{
        \begin{tabular}{c|ccc|ccc}
        \hline
        \multirow{2}{*}{Method} & \multicolumn{3}{c|}{$2cm$} & \multicolumn{3}{c}{$10cm$} \\
        \cline{2-7} & F$_1$ & Comp. & Acc. & F$_1$ & Comp. & Acc. \\   
        \hline  
        \multirow{1}{*}{w/o. Def.} & 87.25 & 86.93 & 87.76 & 96.87 & 96.58 & 97.15 \\ 
        \multirow{1}{*}{w/o. Taj.} & 87.79 & 87.85 & 87.94 & 97.25 & 97.18 & 97.32 \\ 
        \multirow{1}{*}{w/o. Map.} & 88.04 & 88.06 & 88.23 & 97.41 & 97.31 & 97.52 \\ 
        \multirow{1}{*}{w/o. Ref.} & 88.17 & 88.41 & 88.12 & 97.5 & 97.54 & 97.46 \\ 
        \multirow{1}{*}{w/o. Pro.} & 88.29 & 88.19 & 88.57 & 97.57 & 97.36 & 97.79 \\ 
        \hline
        \multirow{1}{*}{w/o. Syn.} & 87.83 & 87.66 & 88.19 & 97.26 & 97.04 & 97.48 \\ 
        \multirow{1}{*}{w/o. Sup.} & 88.26 & 88.29 & 88.43 & 97.57 & 97.43 & 97.71 \\ 
        \multirow{1}{*}{w/o. Ini.} & 88.35 & 88.41 & 88.49 & 97.63 & 97.53 & 97.74 \\ 
        \multirow{1}{*}{w/o. Seg.} & 88.57 & 88.7 & 88.65 & 97.77 & 97.74 & 97.81 \\ 
        \hline
        \multirow{1}{*}{w/o. Occ.} & 88.14 & 88.02 & 88.47 & 97.48 & 97.29 & 97.66 \\ 
        \multirow{1}{*}{w/o. Str.} & 88.31 & 88.3 & 88.52 & 97.6 & 97.48 & 97.72 \\
        \hline
        \multirow{1}{*}{SED-MVS} & \textcolor{red}{\textbf{88.85}} & \textcolor{red}{\textbf{88.94}} & \textcolor{red}{\textbf{88.96}} & \textcolor{red}{\textbf{97.98}} & \textcolor{red}{\textbf{97.92}} & \textcolor{red}{\textbf{98.05}} \\  
        \hline
        \end{tabular}%
    }
    \label{table: ablation study}%
    \vspace{-0.05in}
\end{table}%

\subsubsection{Sparse-Monocular Synergistic Restoration}
We individually remove the entire sparse-monocular synergistic restoration module (w/o. Syn.), restored depth map for supervised guidance (w/o. Sup.) and initialization (w/o. Ini.), and apply segmentation during triangulation when restoring the depth map (w/o. Seg.) for ablations. In comparison, w/o. Syn. produces the lowest F$_1$ score, which emphasizes the necessity of sparse-monocular synergistic restoration. The F$_1$ score of w/o. Ini. slightly exceeds that of w/o. Sup., suggesting that utilizing the restored depth map for supervised guidance during PM iterations is more beneficial for reconstruction than adopting it for initialization before PM iteration, though both contribute significantly. 
The F$_1$ score of w/o. Seg. is second only to SED-MVS, indicating that segmentation during triangulation is somewhat effective in separating different instances.

\subsubsection{Occlusion-Aware Patch Deformation}
We respectively remove the entire occlusion-aware patch deformation module (w/o. Occ.) and strict depth edges constraint (w/o. Str.) during patch deformation for ablations. 
Evidently, w/o. Str performs slightly better than w/o. Occ, but both are inferior to SED-MVS. 
This suggests that excluding strict depth edges constraint during patch deformation is less effective than our proposed dual-categories edge constraint, which incorporates occlusion maps for occlusion-aware patch deformation.

\subsubsection{Ablation Studies on Key Parameters}
Moreover, to justify our parameter choices, we further conduct ablation studies on key parameters, as presented in Table \ref{table: parameters}. 
The best and second-best F$_1$ scores, along with the parameters we finally selected, are highlighted in bold, red, and blue, respectively. 
As indicated, most of the parameters we chose correspond to the best reconstruction results.
Additionally, except for key parameters such as $X$, $\eta$, and $\delta$, most other parameters show low sensitivity. 
For Segmentation-Driven and Edge-Aligned Patch Deformation, although F$_1$ improves when $X$ exceeds 16, it also leads to increased runtime. Therefore, we opt for $X=16$ as a trade-off between time and performance. 
The difference in $w$ becomes negligible when it exceeds 11. 

For Sparse-Monocular Synergistic Restoration, the sensitivity of $\gamma$ and $\kappa$ are lower than that of $\mu$, highlighting the importance of truncation threshold for $E_{dp}$ during supervised guidance. Furthermore, when $\eta$ approaches 0, the EM algorithm may progressively reduce the weight of certain errors, causing a noticeable decline in results.

Regarding Occlusion-Aware Patch Deformation, as $\delta$ decreases, the F$_1$ score becomes similar to those achieved with strict edge constraints. Conversely, as $\delta$ increases, the F$_1$ score tends to resemble those produced by flexible edge constraints. Additionally, both $\sigma$ and $\varepsilon$ exhibit relatively low sensitivity.

\begin{table}
    \centering
    \renewcommand{\arraystretch}{1.07} 
    \caption{Quantitative evaluation of hyperparameter variability on the F$_1$ Score at threshold $2cm$ of ETH3D training dataset.}
    \resizebox{\linewidth}{!}{ 
        \begin{tabular}{c|c|c|c|c|c}
        \hline  
        \multirow{1}{*}{Par.} & \multicolumn{5}{c}{Segmentation-Driven and Edge-Aligned ...} \\
        \hline  
        \multirow{1}{*}{$X$} & 8 & 12 & \textcolor{blue}{\textbf{16}} & 32 & 64 \\ 
        \hline  
        \multirow{1}{*}{F$_1$} & 88.12 & 88.56 & 88.85 & \textbf{88.97} & \textcolor{red}{\textbf{89.06}} \\ 
        \hline  
        \multirow{1}{*}{$w$} & 7 & 9 & \textcolor{blue}{\textbf{11}} & 13 & 15 \\ 
        \hline  
        \multirow{1}{*}{F$_1$} & 88.65 & 88.79 & 88.85 & \textbf{88.88} & \textcolor{red}{\textbf{88.91}} \\ 
        \hline  
        \multirow{1}{*}{Par.} & \multicolumn{5}{c}{Sparse-Monocular Synergistic Restoration} \\
        \hline  
        \multirow{1}{*}{$\gamma$} & 1$\times$10$^{-3}$ & 3$\times$10$^{-3}$ & \textcolor{blue}{\textbf{5$\times$10$^{-3}$}} & 7$\times$10$^{-3}$ & 9$\times$10$^{-3}$ \\ 
        \hline  
        \multirow{1}{*}{F$_1$} & 88.63 & 88.76 & \textcolor{red}{\textbf{88.85}} & \textbf{88.81} & 88.74 \\ 
        \hline  
        \multirow{1}{*}{$\kappa$} & 0.5 & 0.6 & \textcolor{blue}{\textbf{0.7}} & 0.8 & 0.9 \\ 
        \hline  
        \multirow{1}{*}{F$_1$} & 88.69 & \textbf{88.79} & \textcolor{red}{\textbf{88.85}} & 88.71 & 88.54 \\ 
        \hline  
        \multirow{1}{*}{$\mu$} & 0.0125$\times$2$^{n}$ & 0.025$\times$2$^{n}$ & \textcolor{blue}{\textbf{0.05$\times$2$^{n}$}} & 0.1$\times$2$^{n}$ & 0.2$\times$2$^{n}$ \\ 
        \hline  
        \multirow{1}{*}{F$_1$} & 88.49 & 88.70 & \textcolor{red}{\textbf{88.85}} & \textbf{88.75} & 88.61 \\ 
        \hline  
        \multirow{1}{*}{$\eta$} & 0 & 0.05 & \textcolor{blue}{\textbf{0.1}} & 0.15 & 0.2 \\ 
        \hline  
        \multirow{1}{*}{F$_1$} & 88.07 & 88.48 & \textcolor{red}{\textbf{88.85}} & \textbf{88.62} & 88.35 \\ 
        \hline  
        \multirow{1}{*}{Par.} & \multicolumn{5}{c}{Occlusion-Aware Patch Deformation} \\
        \hline  
        \multirow{1}{*}{$\delta$} & 0.6 & 1.2 & \textcolor{blue}{\textbf{1.8}} & 2.4 & 3.0 \\ 
        \hline  
        \multirow{1}{*}{F$_1$} & 88.39 & 88.64 & \textcolor{red}{\textbf{88.85}} & \textbf{88.72} & 88.52 \\ 
        \hline  
        \multirow{1}{*}{$\sigma$} & 4 & 8 & \textcolor{blue}{\textbf{12}} & 24 & 36 \\ 
        \hline  
        \multirow{1}{*}{F$_1$} & 88.74 & 88.79 & \textcolor{red}{\textbf{88.85}} & \textbf{88.83} & 88.79 \\ 
        \hline  
        \multirow{1}{*}{$\varepsilon$} & 2$\times$2$^{n}$ & 4$\times$2$^{n}$ & \textcolor{blue}{\textbf{8$\times$2$^{n}$}} & 16$\times$2$^{n}$ & 32$\times$2$^{n}$ \\
        \hline  
        \multirow{1}{*}{F$_1$} & 88.69 & 88.78 & \textcolor{red}{\textbf{88.85}} & \textbf{88.82} & 88.81 \\ 
        \hline  
        \end{tabular}%
    }
    \label{table: parameters}%
    \vspace{-0.05in}
\end{table}%

\section{Conclusion}
In this paper, we propose SED-MVS, which leverages panoptic segmentation and multi-trajectory diffusion strategy for segmentation-driven and edge-aligned patch deformation. 
To prevent edge-skipping, we first employ panoptic segmentation to extract depth edge as guidance for patch deformation, and further employ a multi-trajectory diffusion strategy to enable deformed patch with edge-alignment capacity.
Moreover, to avoid inaccuracy caused by random initialization, we combine sparse points and monocular depth map to restore reliable depth map for initialization and supervised guidance.
Additionally, we integrate segmentation image with monocular depth map to generate occlusion map for dual-categories edge constraint, thus enabling occlusion-aware patch deformation.
Evaluation results on ETH3D, Tanks \& Temples, BlendedMVS and Strecha datasets validate the superior 
performance and robust generalization capability of our proposed method. 
However, the algorithm still faces challenges when dealing with large textureless curved surfaces. Future work tends to focus on exploring curved surface assumptions for patch deformations to address this limitation.

\bibliography{SED-MVS}
\bibliographystyle{IEEEtran}

\begin{IEEEbiography}[{\includegraphics[width=1in,height=1.3in,clip,keepaspectratio]{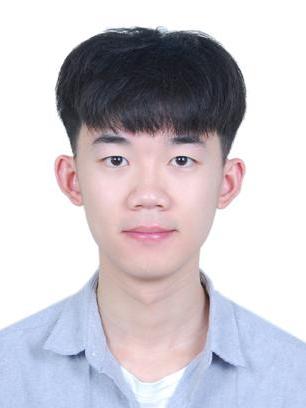}}]{Zhenlong Yuan}
received the B.Sc. degree in telecommunications engineering with management from Beijing University of Posts and Telecommunications, Beijing, China. He is working toward the Ph.D. degree in the Institute of Computing Technology, Chinese Academy of Sciences and University of Chinese Academy of Sciences. His main research interests include 3D reconstruction.
\end{IEEEbiography}

\begin{IEEEbiography}[{\includegraphics[width=1in,height=1.3in,clip,keepaspectratio]{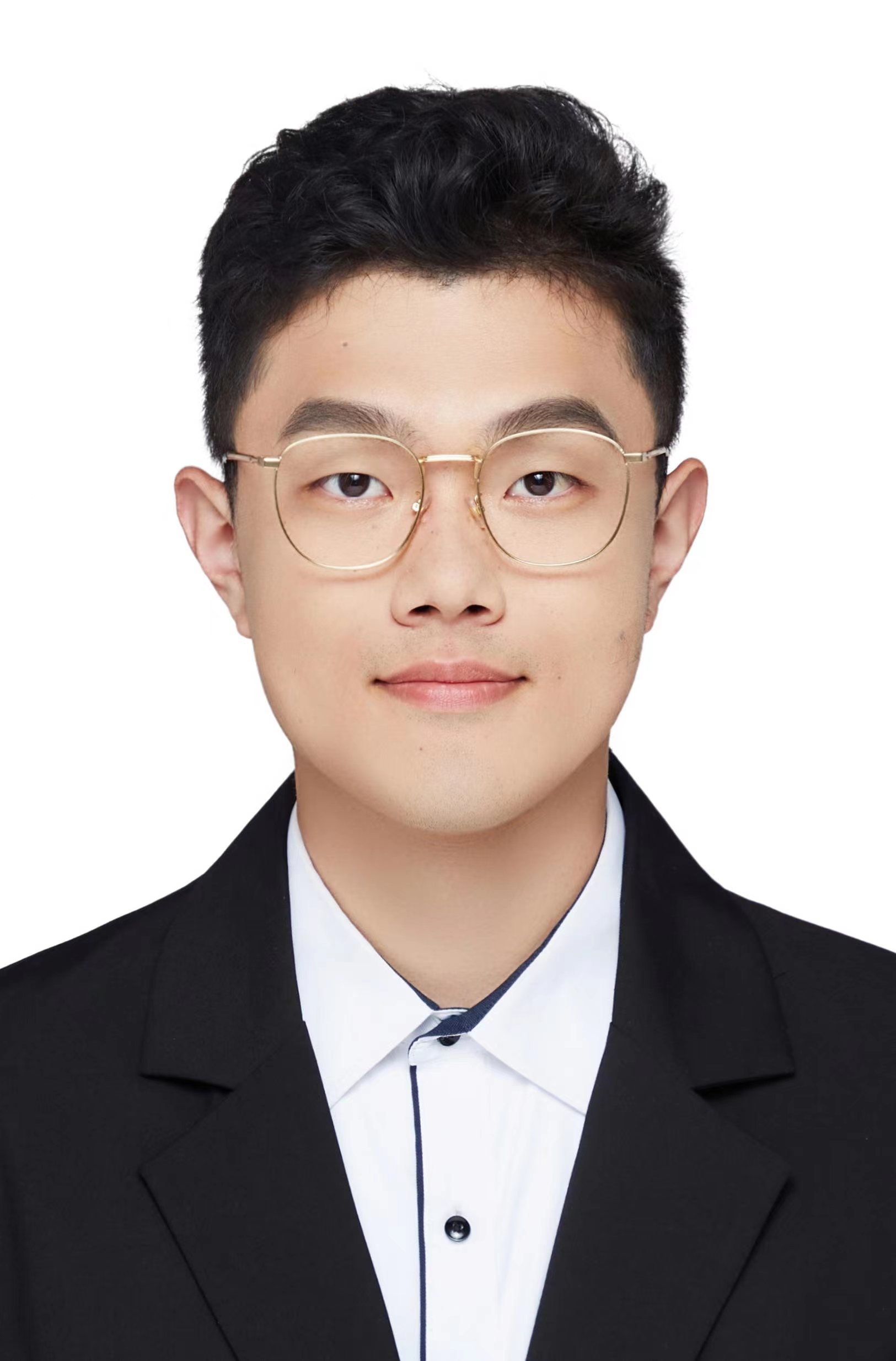}}]{Zhidong Yang}
received the B.Sc. degree in digital media technology from Northeastern University, Shenyang, China. He is working towards the Ph.D. degree at the Institute of Computing Technology, Chinese Academy of Sciences and University of Chinese Academy of Sciences. His main research interests include image restoration, biomedical image analysis, and 3D reconstruction.
\end{IEEEbiography}

\begin{IEEEbiography}[{\includegraphics[width=1in,height=1.3in,clip,keepaspectratio]{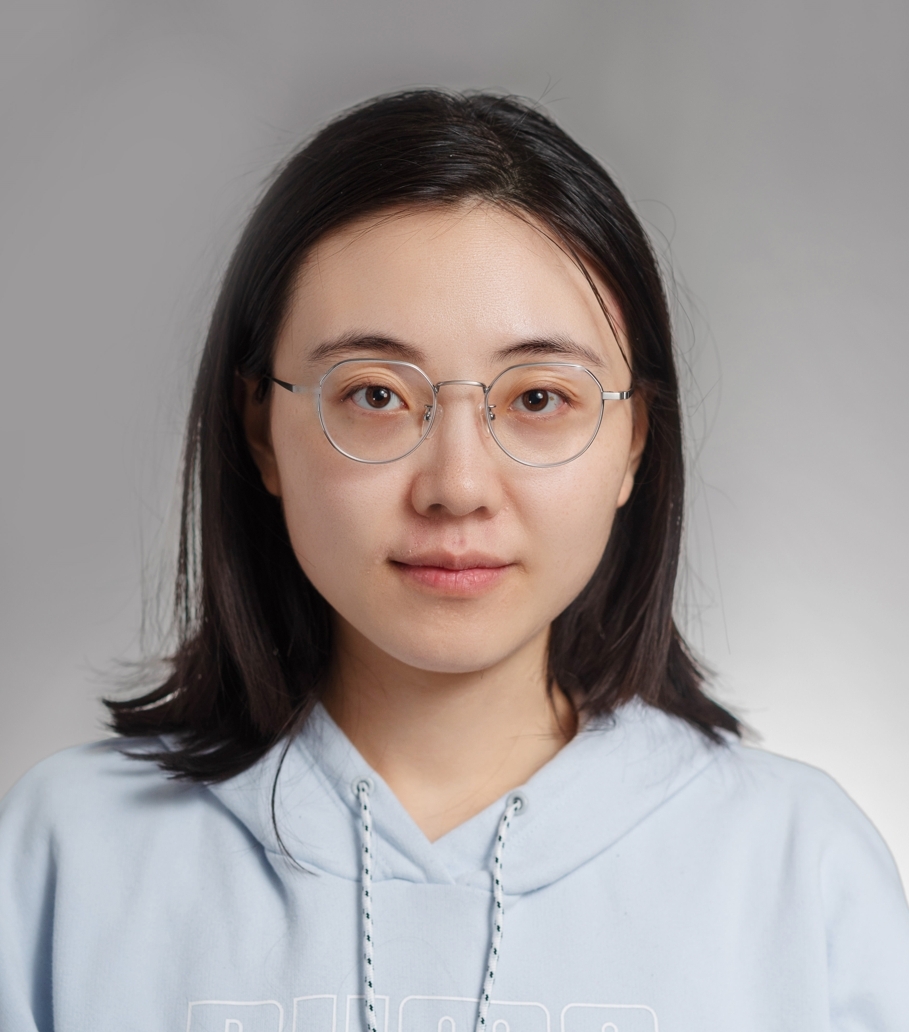}}]{Yujun Cai}
is currently a Lecturer at the School of Electrical Engineering and Computer Science, the University of Queensland (UQ). Before that she was a Research Scientist in Meta Inc. She obtained her Ph.D. from Nanyang Technological University in 2021. She served as Area Chair for ICML, WWW, ICCV, IJCAI, WACV etc. She’s also reviewer of IEEE Trans. on Pattern Analysis and Machine Intelligence(T-PAMI), International Journal of Computer Vision (IJCV), IEEE Trans. on Image Processing (T-IP), IEEE Trans. on Circuits and Systems for Video Technology (T-CSVT) etc.
\end{IEEEbiography}

\begin{IEEEbiography}[{\includegraphics[width=1in,height=1.3in,clip,keepaspectratio]{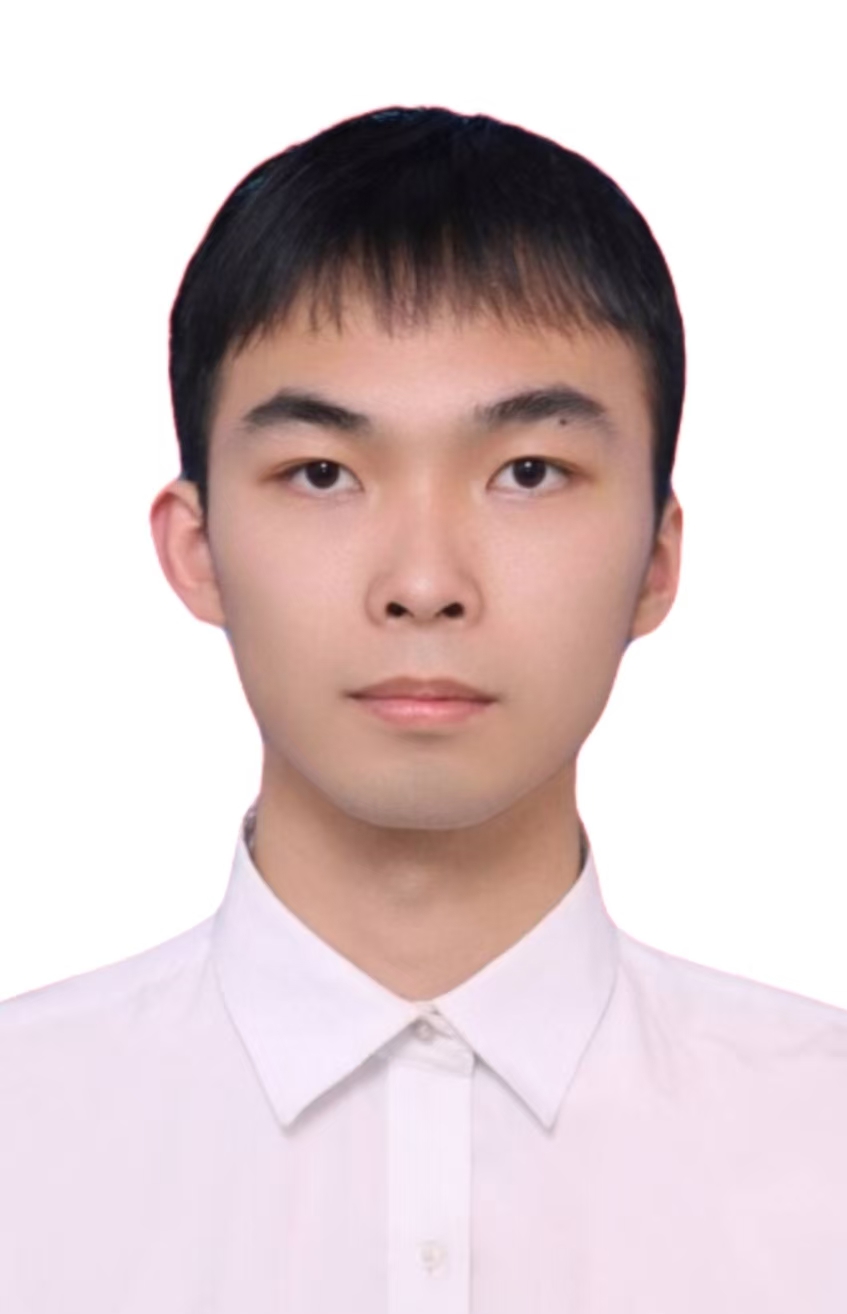}}]{Kuangxin Wu}
received the B.Sc. degree in Cybersecurity and Law Enforcement from Hunan Police Academy, Changsha, China. His research focuses on autonomous driving, vision language model, and 3D reconstruction.
\end{IEEEbiography}

\begin{IEEEbiography}[{\includegraphics[width=1in,height=1.3in,clip,keepaspectratio]{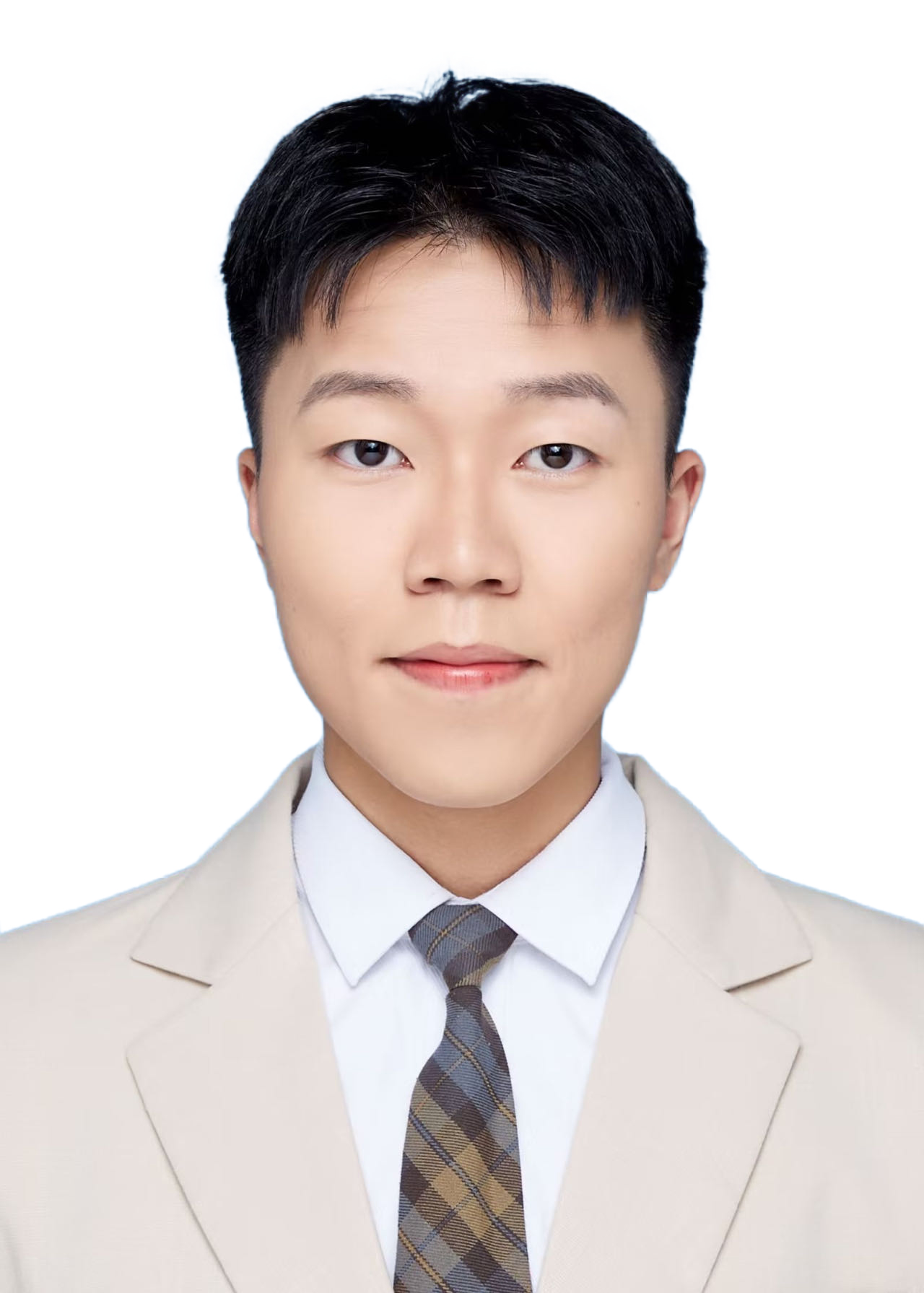}}]{Mufan Liu}
received the B.Sc. degree in communication engineering from the University of Electronic Science and Technology of China, Chengdu, China, in 2023. He is currently pursuing the Ph.D. degree with the Cooperative MediaNet Innovation Center, Shanghai Jiao Tong University, Shanghai, China. His main research interests include adaptive streaming, joint source and channel coding and multimedia processing.
\end{IEEEbiography}

\begin{IEEEbiography}[{\includegraphics[width=1in,height=1.3in,clip,keepaspectratio]{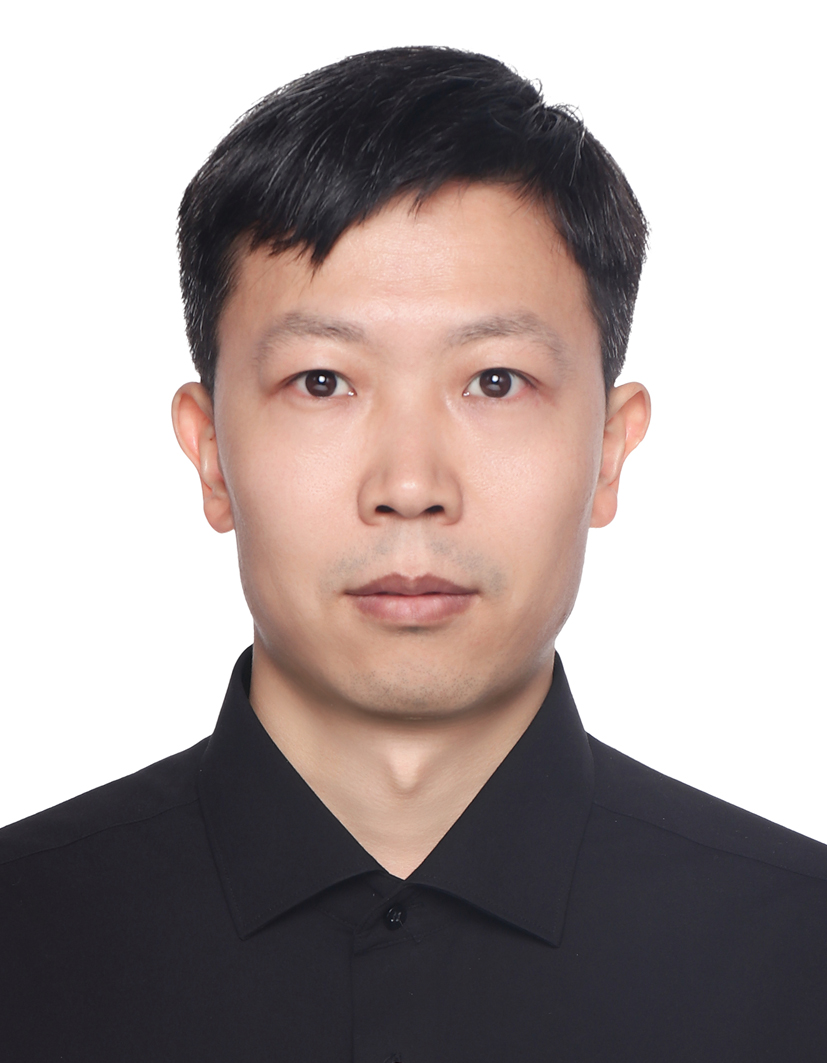}}]{Dapeng Zhang}
is a Ph.D. candidate in the School of Information Science and Engineering at Lanzhou University. His research focuses on autonomous driving and world model.
\end{IEEEbiography}

\begin{IEEEbiography}[{\includegraphics[width=1in,height=1.3in,clip,keepaspectratio]{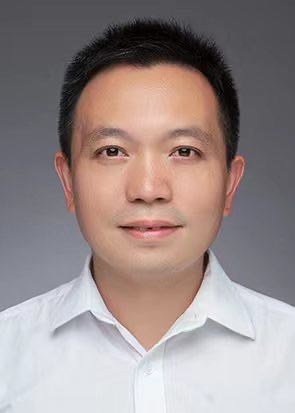}}]{Hao Jiang}
obtained Ph.D from the Institute of Computing Technology, Chinese Academy of Sciences. Currently, he is an associate professor at the Institute of Computing Technology, Chinese Academy of Sciences and University of Chinese Academy of Sciences. His research interests include virtual reality/augmented reality, computer graphics, and intelligent user interfaces.
\end{IEEEbiography}

\begin{IEEEbiography}[{\includegraphics[width=1in,height=1.3in,clip,keepaspectratio]{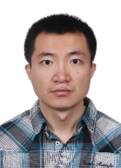}}]{Zhaoxin Li}
received the Ph.D. degree in computer application technology from Harbin Institute of Technology, Harbin, China, in 2016. From July 2016 to April 2023, he worked as an Assistant Professor in the Institute of Computing Technology, Chinese Academy of Sciences. From September 2018 to March 2019, he worked as a Postdoctoral Fellow in the Department of Computing, The Hong Kong Polytechnic University. He is currently with the Agricultural Information Institute of Chinese Academy of Agricultural Sciences, China. His research interests include 3D computer vision and 3D data processing.
\end{IEEEbiography}

\begin{IEEEbiography}[{\includegraphics[width=1in,height=1.3in,clip,keepaspectratio]{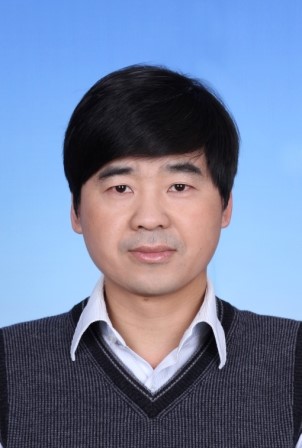}}]{Zhaoqi Wang}
is a researcher and a director of PhD students with the Institute of Computing Technology, Chinese Academy of Sciences. His research interests include virtual reality and intel-ligent human computer interaction. He is a senior member of the China Computer Federation.
\end{IEEEbiography}

\vfill

\end{document}